\documentclass[10pt,journal,compsoc]{IEEEtran}
\ifCLASSOPTIONcompsoc
\usepackage[nocompress]{cite}
\else
\usepackage{cite}
\fi
\ifCLASSINFOpdf
\else
\fi

\usepackage{amssymb}

\usepackage{amsfonts}

\usepackage{bbm}
\usepackage{bm}
\usepackage{color}
\usepackage{cite}
\usepackage[linesnumbered,ruled]{algorithm2e}
\usepackage{ulem} 
\usepackage{hyperref}
\usepackage{float}
\usepackage[T1]{fontenc}
\usepackage{booktabs}
\usepackage{multirow}
\usepackage{mathrsfs}
\usepackage[utf8]{inputenc}
\usepackage{ulem}
\usepackage{color}
\usepackage[table]{xcolor}
\usepackage{float}

\newcounter{RNum}

\usepackage{graphics} 
\usepackage{epsfig} 
\usepackage{times} 
\usepackage{amsmath} 

\usepackage{amsfonts}

\usepackage{wrapfig}
\usepackage{graphicx}
\usepackage{pifont}
\usepackage{color}
\usepackage{hyperref}
\usepackage{soul} 
\DeclareSymbolFont{largesymbol}{OMX}{yhex}{m}{n}
\DeclareMathAccent{\Widehat}{\mathord}{largesymbol}{"62}

\usepackage{graphics} 
\usepackage{epsfig} 
\usepackage{times} 
\usepackage{amsmath} 

\usepackage[caption=false,font=footnotesize]{subfig}

\usepackage{amsfonts}

\usepackage{graphicx}
\usepackage{pifont}

\soulregister\ref7
\soulregister\cite7
\soulregister\url7

\usepackage{booktabs}
\usepackage{multirow}
\usepackage{pifont}

\usepackage{ragged2e}
\usepackage{bbding}

\begin{document}
	%
	\title{DiffTF++: 3D-aware Diffusion Transformer for Large-Vocabulary 3D Generation}
	%
	%
	%
	%
	
	\author{Ziang~Cao,~\IEEEmembership{Student Member,~IEEE,}
		Fangzhou~Hong,~\IEEEmembership{Student Member,~IEEE,}
		Tong~Wu,~\IEEEmembership{Student Member,~IEEE,}
		Liang~Pan,~\IEEEmembership{Member,~IEEE,}
		and~Ziwei~Liu,~\IEEEmembership{Member,~IEEE}
		\IEEEcompsocitemizethanks{\IEEEcompsocthanksitem Z. Cao, F. Hong, and Z. Liu are with the School of Computer Science and Engineering, S-Lab, Nanyang Technological University, Singapore. \protect\\
			E-mail: {ziang.cao, fangzhou.hong, ziwei.liu}@ntu.edu.sg

   \IEEEcompsocthanksitem L. Pan is with the Shanghai Artificial Intelligent Laboratory, China. \protect\\
			E-mail: liang.pan@ntu.edu.sg
			\IEEEcompsocthanksitem T. Wu is with the graduate division of Information Engineering, The Chinese University of Hong Kong, China. \protect\\
			E-mail: wt020@ie.cuhk.edu.hk

		}
		
		\thanks{Manuscript received April 1, 2024.}
	}
	
	%
	%

	\markboth{IEEE TRANSACTION ON PATTERN ANALYSIS AND MACHINE INTELLGENCE}%
	{Shell \MakeLowercase{\textit{et al.}}: Bare Demo of IEEEtran.cls for IEEE Journals}
	%



	\IEEEtitleabstractindextext{%
		\begin{abstract}
			\justifying
			Generating diverse and high-quality 3D assets automatically poses a fundamental yet challenging task in 3D computer vision. Despite extensive efforts in 3D generation, existing optimization-based approaches struggle to produce large-scale 3D assets efficiently. Meanwhile, feed-forward methods often focus on generating only a single category or a few categories, limiting their generalizability. Therefore, we introduce a diffusion-based feed-forward framework to address these challenges \textit{with a single model}. To handle the large diversity and complexity in geometry and texture across categories efficiently, we \textbf{1}) adopt improved triplane to guarantee efficiency; \textbf{2}) introduce the 3D-aware transformer to aggregate the generalized 3D knowledge with specialized 3D features; and \textbf{3}) devise the 3D-aware encoder/decoder to enhance the generalized
3D knowledge. Building upon our 3D-aware \textbf{Diff}usion model with \textbf{T}rans\textbf{F}ormer, \textbf{DiffTF}, we propose a stronger version for 3D generation, \textit{i.e.}, \textbf{DiffTF++}. It boils down to two parts: multi-view reconstruction loss and triplane refinement. 
        Specifically, we utilize multi-view reconstruction loss to fine-tune the diffusion model and triplane decoder, thereby avoiding the negative influence caused by reconstruction errors and improving texture synthesis. By eliminating the mismatch between the two stages, the generative performance is enhanced, especially in texture. Additionally, a 3D-aware refinement process is introduced to filter out artifacts and refine triplanes, resulting in the generation of more intricate and reasonable details. Extensive experiments on ShapeNet and OmniObject3D convincingly demonstrate the effectiveness of our proposed modules and the state-of-the-art 3D object generation performance with large diversity, rich semantics, and high quality.

		\end{abstract}
		
		}

	\maketitle

	\IEEEdisplaynontitleabstractindextext

	%
	\IEEEpeerreviewmaketitle

	\IEEEraisesectionheading{\section{Introduction}\label{sec:introduction}}

	%
	%
	%
	%
	\IEEEPARstart{A}{s} one of the fundamental tasks in 3D computer vision, generating diverse and high-quality 3D content has experienced rapid developments. These 3D assets play a pivotal role in fostering the progress of numerous applications, such as constructing game scenes, robotics, and architecture. However, large-vocabulary 3D object generation still has several special challenges: a) the urgent requirement for expressive yet efficient 3D representation; b) wide diversity across massive categories; and c) the complicated appearance of real-world objects.

While several advanced techniques have demonstrated remarkable achievements in 3D generation, optimized-based methods~\cite{poole2022dreamfusion,Lin_2023_CVPR,wang2023score, wang2024prolificdreamer} often suffer from inefficiency. Generally, these methods require time-consuming optimization processes tailored to each individual 3D object, significantly hampering efficiency and feasibility. Besides, optimized-based methods struggle to harness the potential benefits of large-scale datasets. On the other hand, feed-forward models~\cite{sitzmann2019deepvoxels,muller2022diffrf,achlioptas2018learning,nichol2022point,chan2022efficient,mo2023dit} offer a promising alternative by leveraging large-scale data to enhance their generalization capabilities. However, most existing feed-forward models are constrained to a single or a few categories due to weak 3D awareness, hindering their ability to generalize across diverse 3D datasets.

 

 To tackle the challenges, we propose a 3D-aware coarse-to-fine framework, \textit{i.e.}, \textbf{DiffTF++} for general 3D object generation. 
 Firstly, we address the trade-off between computational intensity and extendibility by adopting a hybrid 3D representation known as the Triplane feature~\cite{chan2022efficient}. Moreover, to effectively handle the diverse and complex appearance of real-world objects, we conceptualize objects as compositions of generalized 3D prior knowledge and specialized 3D features. This approach allows us to leverage large-vocabulary objects while reducing generation complexity.  By employing a specialized share-weight cross-plane attention module, our transformer facilitates the generalization of 3D prior knowledge across different object categories. Additionally, by applying distinct self-attention mechanisms to specialized 3D features, our transformer effectively integrates 3D interdependencies, thereby enhancing performance in handling diverse objects. Furthermore, to enhance 3D awareness and semantic understanding in triplanes for intricate categories, we develop a 3D-aware encoder/decoder architecture. Unlike traditional 2D encoders that may compromise 3D relations, our model adaptively integrates 3D awareness into encoded features, thereby enriching the representation of 3D-related information.


\begin{figure*}[t]
		\centering

		\includegraphics[width=1\textwidth]{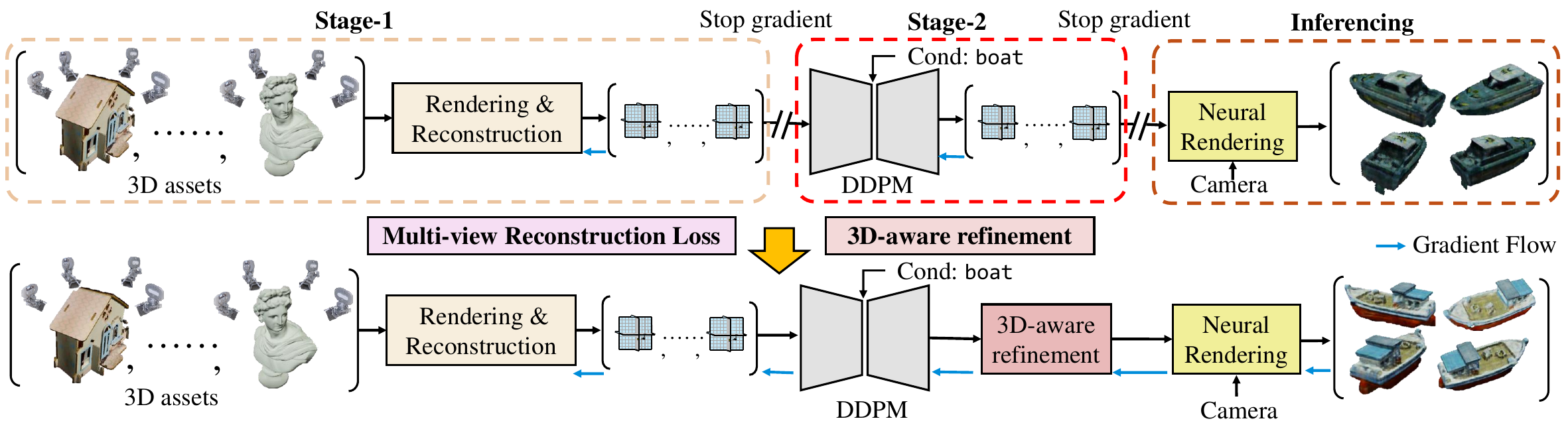} 
		
		\caption{\textbf{Comparison between DiffTF (Top) and DiffTF++ (Bottom).} By introducing multi-view reconstruction loss and 3D-aware refinement, DiffTF++ can not only improve the quality of texture but reduce the artifacts in the generated 3D objects. }\label{fig:framework}
  \includegraphics[width=1\textwidth]{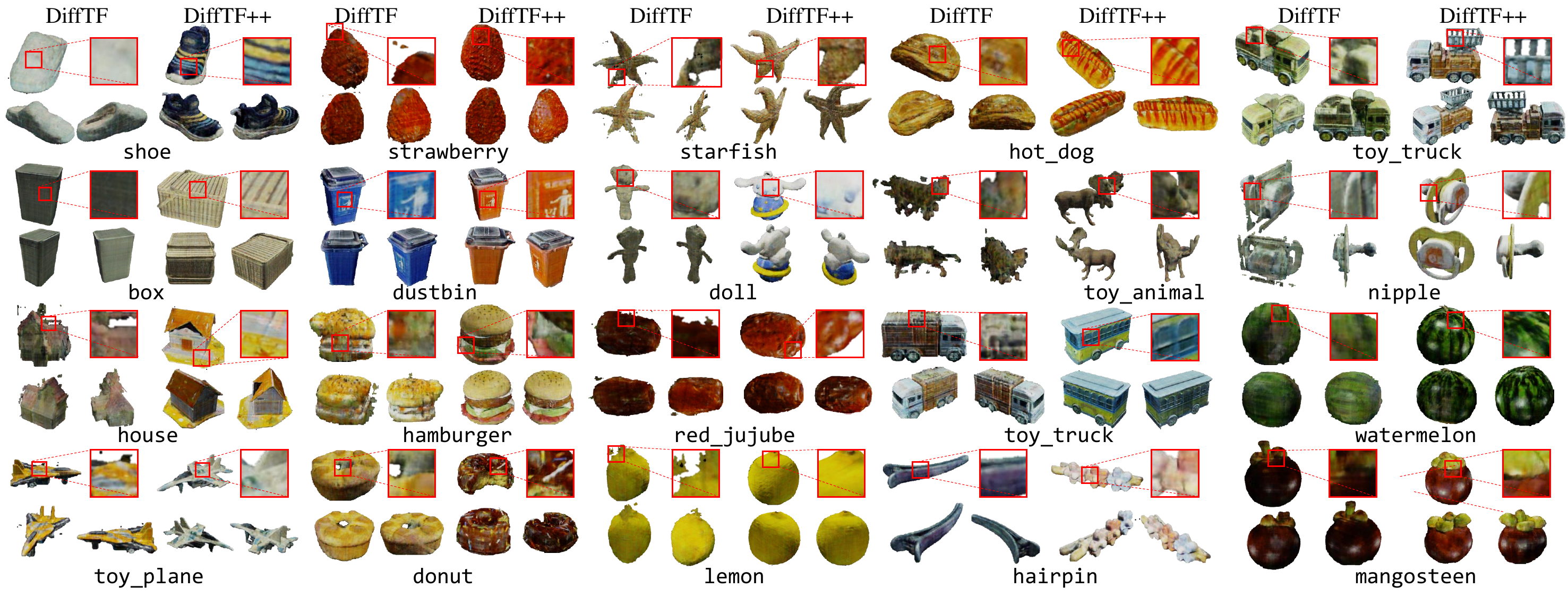}
  \caption{\textbf{Visualization on Large-vocabulary 3D object generation.} By introducing 3D-aware refinement and eliminating the mismatch between the two stages, DiffTF++ not only reduces the artifacts in 3D objects but improves the diversity and accuracy of the generated 3D assets in terms of texture and topology.}\label{fig:vis}

	\end{figure*}
 
 In addition, to achieve even higher performance, we introduce a multi-view reconstruction loss to establish a connection between the two stages, \textit{i.e.}, triplane fitting and diffusion training depicted in Fig.~\ref{fig:framework}. This loss function allows our model to eliminate the discrepancy between the ground truth and the reconstructed triplane in the first stage by backpropagating gradients. Therefore, it can enhance the generative performance, especially in texture. Concurrently, we incorporate a 3D-aware refinement mechanism, comprising a cross-plane attention module, to eliminate artifacts and enhance the quality of the generated triplane. The integration of the multi-view reconstruction loss and 3D-aware refinement significantly enhances the capabilities of DiffTF++, enabling it to generate more realistic 3D assets with impressive texture and topology. As demonstrated in Fig.~\ref{fig:vis}, our proposed approach effectively eliminates artifacts and enriches the detailed topology and texture of the generated 3D objects. Compared to the previous version (DiffTF), DiffTF++ excels in generating intricate and compelling 3D content with rich semantics and realistic topology and texture details. To validate the effectiveness of our proposed method in large-vocabulary 3D object generation, we conduct exhaustive experiments on two widely recognized benchmarks, \textit{i.e.}, ShapeNet~\cite{chang2015shapenet} and OmniObject3D~\cite{wu2023omniobject3d}. Notably, OmniObject3D comprises 216 intricate categories, encompassing objects with complex geometry and texture such as pine cones, pitaya, and various vegetables. This diversity presents a rigorous testbed for assessing the generalization capability of our approach. We employed both 2D and 3D metrics on these benchmarks to quantitatively and qualitatively evaluate the generative performance and effectiveness of our proposed modules. Additionally, we conducted a comprehensive user study to gauge human preference for rendered images, providing further convincing validation to our performance evaluation. The results across all metrics strongly demonstrate the superior generative performance of our method compared to existing approaches in generating high-quality 3D objects across diverse categories.

	 \begin{figure*}[t]
		\centering

		\includegraphics[width=1\textwidth]{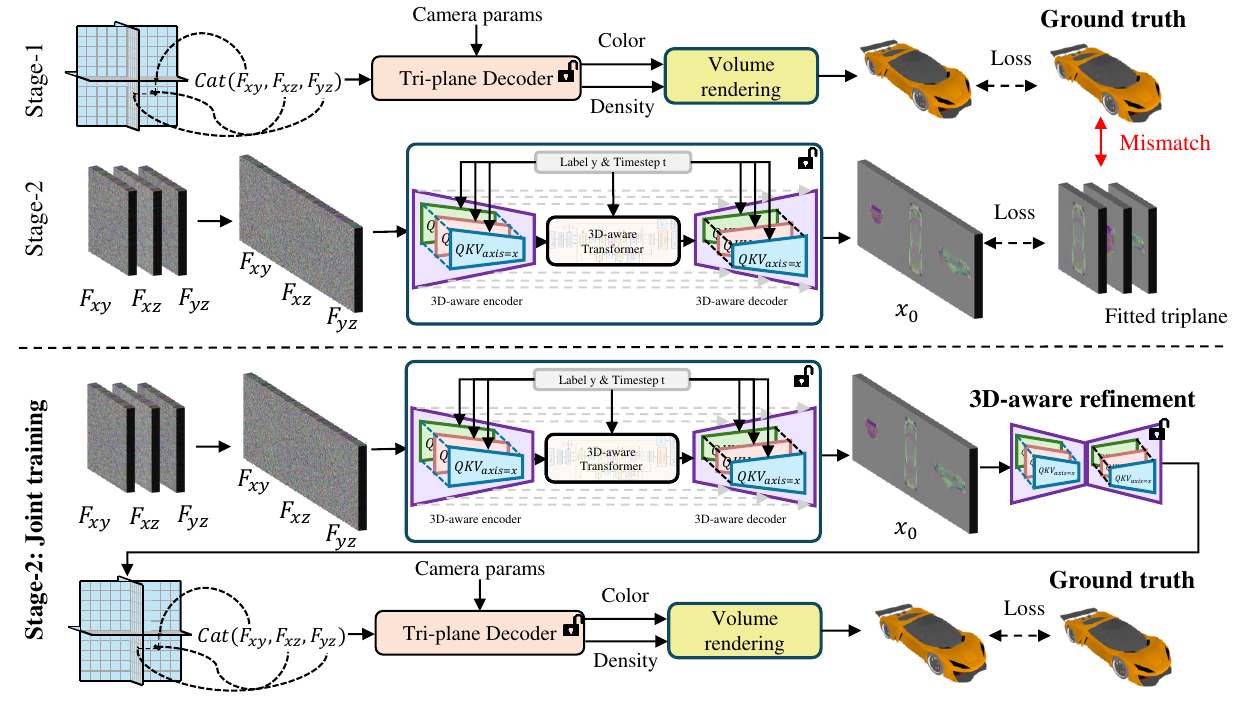}
		
		\caption{\textbf{Pipeline comparison between DiffTF and DiffTF++.}  DiffTF (top) has two individual stages: 1) optimize the triplane features for each 3D object, and 2) train our 3D-aware diffusion model based on those fitted triplanes. Due to the gradient discontinuity between the two stages, there exists an inevitable mismatch between the objectives of diffusion and the ground truth. To handle it, we adopt a multi-view reconstruction loss (bottom) for DiffTF++ by introducing a neat 3D-aware refinement and a multi-view reconstruction loss.}
		\label{fig:conv}

	\end{figure*}
	
	This paper is an extension of our conference version accepted to ICLR2024~\cite{cao2023large}. Compared with the earlier version, \textit{i.e.}, DiffTF, we provide novel and valuable materials including a 3D-aware refinement, a multi-view reconstruction loss to overcome the mismatch between the two stages, exhaustive experiments, additional comprehensive user studies, and detailed implementations that are clearly summarized as follows:
	
	\begin{itemize}
	
		\item To enhance the overall quality of the generated triplane and eliminate artifacts, we introduce a 3D-aware refinement mechanism based on our novel cross-plane attention approach. It can effectively filter out noise information, leading to a substantial improvement in the quality of the generated objects.

  
  
  \item To address the mismatch between two independent stages and promote better convergence, we incorporate a multi-view reconstruction loss. It allows our framework to backpropagate gradients across stages, eliminating the effects of the mismatch and promoting the convergence performance of the diffusion model.
  

		\item We conduct comprehensive experiments on two benchmarks, proving the state-of-the-art (SOTA) generative performance on large-vocabulary generation. In addition to traditional 2D and 3D metrics, we also considered human preferences. It demonstrates that our generated 3D assets are visually appealing and perceptually aligned with human standards.

		\item On top of the conference version, we provide more analysis of existing generative methods, detailed exposition of the proposed structure, information about implementation, and additional experiment results.
	
	\end{itemize}
	

		
		

	\section{Related Work}
	
	In this section, we introduce the recent related improvements in 3D generative models including GAN-based and diffusion-based methods, as well as transformer structure.
	
	\subsection{Transformer} 
	In recent years, Transformer~\cite{vaswani2017attention} has seen rapid progress in many fields including image recognition~\cite{dosovitskiyimage,touvron2021training}, object detection~\cite{carion2020end,zhu2020deformable}, tracking~\cite{cao2021hift,cao2022tctrack,cao2023towards}, segmentation~\cite{zheng2021rethinking,strudel2021segmenter}, and image generation~\cite{van2016conditional,jiang2021transgan,mo2023dit}.  Some works~\cite{chen2020generative,child2019generating} prove the remarkable of transformer when predicting the pixels autoregressively. Based on the masked token, MaskGIT.~\cite{chang2022maskgit} achieve promising generation performance. DiT~\cite{peebles2022scalable} adopts the transformer as the backbone of diffusion models of images. Based on the 2D version, \cite{mo2023dit} propose a DiT-3D for point cloud generation.  Despite impressive progress in transformer-based generative models, they are optimized on a single or a few categories. In this paper, we propose a 3D-aware transformer for diverse real-world 3D object generation that can extract the generalized 3D prior knowledge and specialized 3D features from various categories.
	
	\begin{figure}[t]
		\centering

		\includegraphics[width=0.5\textwidth]{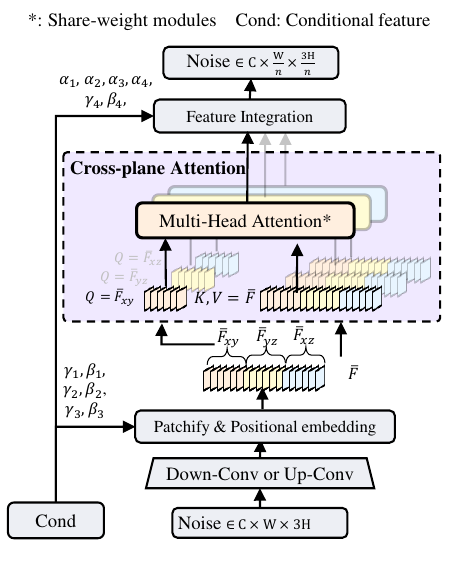}
		
		\caption{\textbf{The detailed structure of our proposed 3D-aware encoder/decoder}. The 3D-aware module can efficiently encode the triplanes while maintaining the 3D-related information via a single cross-plane attention module.}
		\label{fig:details}
		
	\end{figure}
	
	\subsection{3D Generation} 
	With satisfying performance in 2D image synthesis, generative adversarial networks (GANs) inspire research in 3D via generating meshes~\cite{gao2022get3d}, texture~\cite{siddiqui2022texturify}, voxel~\cite{chen2019text2shape}, NeRFs~\cite{chan2021pi,gu2021stylenerf,niemeyer2021giraffe,schwarz2020graf,zhou2021cips}, Triplane~\cite{chan2022efficient}, and point cloud~\cite{achlioptas2018learning}. Since the NeRFs and Triplane-based methods can adopt 2D images as supervision without any 3D assets, those two branches of methods have received much attention recently. By adding a standard NeRF model in the GAN model, Pi-GAN~\cite{chan2021pi} and GRAF~\cite{schwarz2020graf} can generate novel view synthesis of the generated 3D objects. Since the high memory loading and long training time of NeRF, adopting high-resolution images is hard to afford which impedes the generative performance. GIRAFFE~\cite{niemeyer2021giraffe} propose a downsample-upsample structure to handle this problem. By generating low-resolution feature maps and upsampling the feature, GIRAFFE~\cite{niemeyer2021giraffe} indeed improves the quality and resolution of output images. To address the 3D inconsistencies, StyleNeRF~\cite{gu2021stylenerf} design a convolutional stage to minimize the inconsistencies. To boost the training efficiency further, EG3D~\cite{chan2022efficient} propose an efficient 3D representation, \textit{i.e.}, triplane. Due to its promising efficiency, in this work, we adopt the revised triplane as our 3D representation.
	
	In contrast to GANs, diffusion models are relatively unexplored tools for 3D generation. A few valuable works based on NeRF~\cite{poole2022dreamfusion,li20223ddesigner}, point cloud~\cite{nichol2022point,luo2021diffusion,zeng2022lion}, triplane~\cite{shue20223d,wang2022rodin,gu2023nerfdiff} show the huge potential of diffusion model. 
 
 \textbf{Optimized-based methods.} As one of the most representative optimized-based methods, DeamFusion~\cite{poole2022dreamfusion} presents a method to gain NeRFs data and apply the generation based on the pre-train 2D text-image diffusion model. By optimizing NeRFs using 2D diffusion prior, DeamFusion can obtain a promising performance. Similarly, SJC~\cite{wang2023score} applies the chain rule on the learned gradients, and back-propagate the score of a 2D pre-trained diffusion model through the Jacobian of a differentiable renderer. Based on them, Magic3D~\cite{Lin_2023_CVPR} adopt a coarse-to-fine framework to improve the performance and accelerate the convergence. To improve the quality and diversity of generated objects, ProlificDreamer~\cite{wang2024prolificdreamer} adopts Variational Score Distillation. Despite the impressive performance, those optimized-based methods are generally time-consuming and cannot utilize large-scale training data fully. 
 
 \textbf{Feed-forward methods.} Different from them, feed-forward methods are efficient. Since they are generally trained on large-scale data, they are superior in robustness and feasibility. NFD~\cite{shue20223d} adopts a feed-forward framework and views the triplane as the flattened 2D features and utilizes the 2D diffusion. It is indeed that adopting pre-train 2D diffusion can accelerate the training process. However, 2D prior knowledge also limits the capacity of the model in 3D-related information. To handle this problem, Rodin~\cite{wang2022rodin} proposes a 3D-aware convolution module in diffusion. Based on local CNN-based awareness, it can provide local 3D-related information for single-category generation. However, it is hard to maintain robustness when facing large categories for local 3D awareness. To this end, in this paper, we try to utilize global 3D awareness in our diffusion-based feed-forward model to extract the 3D interdependencies across the planes and enhance the relations within individual planes. Besides, to avoid the mismatch between the two independent stages, we adopt a multi-view reconstruction loss in our new method. Furthermore, we introduce a 3D-aware refinement based on our cross-plane attention module to eliminate the artifacts and refine the detailed information in generated 3D assets.

	
	\begin{figure*}[t]
		\centering

		\includegraphics[width=0.9\textwidth]{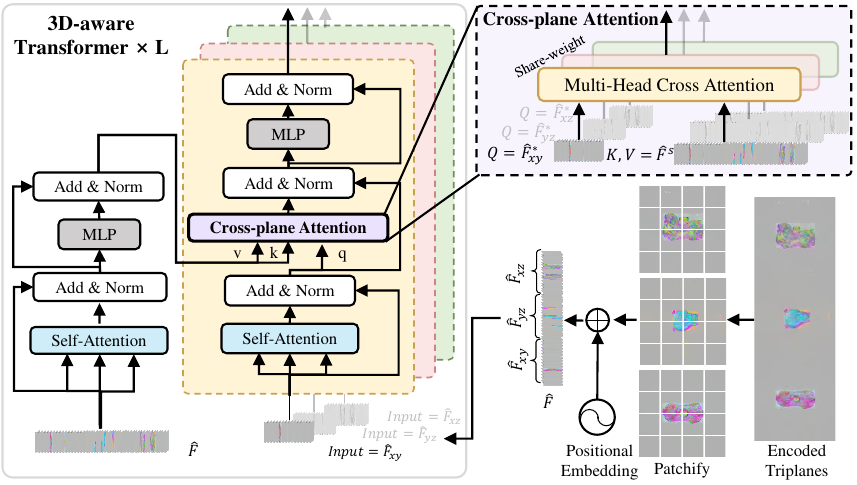}
		
		\caption{\textbf{The detailed structure of our proposed 3D-aware transformer modules.} We take the feature from the xy plane $\hat{F}_{xy}$ as an example. Relying on the extracted generalizable 3D knowledge and specialized one, our model can achieve impressive adaptivity among various categories. }
		\label{fig:detail}
		
	\end{figure*}
	
	\section{Revisit Multi-Head Attention and DDPMs }
	
	\textbf{Multi-Head Attention:} Multi-Head Attention is a fundamental component in transformer structure~\cite{vaswani2017attention} which can be formulated by:
	\begin{equation}\label{att}
		\begin{aligned}
			&\mathrm{MultiHead}({Q},{K},{V})=\Big(\mathrm{Cat}({H}_{att}^1,...,{H}_{att}^N)\Big){W}\\
			&{H}_{att}^{n}=\mathrm{Attention}({Q}{W}^n_q,{K}{W}_k^n,{V}{W}_v^n)\\
			&\mathrm{Attention}({Q},{K},{V})=\mathrm{Softmax}(\dfrac{{Q}{K}^{\mathrm{T}}}{\sqrt{d}}){V}\\
		\end{aligned}
		~ ,
	\end{equation}
	where ${Q}, {K},$ and ${V}$ represent the query, key, and value in the attention operation, ${W}\in~\mathbb{R}^{C_i\times C_i}$, ${W}^n_q\in~\mathbb{R}^{C_i\times C_h}$, ${W}^n_k\in~\mathbb{R}^{C_i\times C_h}$, and ${W}^n_v\in~\mathbb{R}^{C_i\times C_h}$ are learnable weights, and $\sqrt{d}$ is the scaling factor to avoid gradient vanishing.

	\textbf{DDPM:} To solve the generation problem, denoising diffusion probabilistic models~\cite{ho2020denoising} (DDPMs) define the forward and reverse process which transfer the generation problem into predicting noise. The forward process represents the process of applying noise to real data $x_0$ as $q(x_t|x_{t-1})=\mathcal{N}(x_t;\sqrt{1-\beta_t}x_{t-1},\beta_t\mathbf{I})$, where $\beta_t$ and $\mathbf{I}$ represents the forward process variances and unit matrix. For clarification, we assume there are $T$ steps in the forward process. Thus, the features with different noised level can be denoted as $x_{T},x_{T-1},...,x_0$, where $x_{T}$ is sampled from a standard Gaussian noise. Based on the relationship between two continuous step, we have $x_t(x_0,\epsilon)=\sqrt{\overline{\alpha}}_tx_0+\sqrt{1-\overline{\alpha}_t}\epsilon$, where $\epsilon \sim \mathcal{N}(0,\mathbf{I}),\alpha_t=1-\beta_t$ and $\overline{\alpha}_t=\prod_{i=1}^{t}\alpha_i$
	
	During the reverse process, the diffusion model aims to predict the $q(x_{t-1}|x_{t})$. Based on Bayes' theorem and specific parameterization, the $q(x_{t-1}|x_{t})$ can be formulated as:
	$q(x_{t-1}|x_{t})=\mathcal{N}(x_{t-1};\tilde{\mu}_t(x_t,t),\tilde{\beta}_t\mathbf{I})$, where $\tilde{\mu}_t(x_t,t)=\dfrac{1}{\sqrt{\alpha_t}}(x_t-\dfrac{\beta_t}{\sqrt{1-\overline{\alpha}_t}}\epsilon)$ and $\tilde{\beta}_t=\dfrac{1-\overline{\alpha}_{t-1}}{1-\overline{\alpha}_t}\beta_t$. In the end, the objective of reverse process is transferred to predict $\epsilon$. Thus the objective of training is minimize the loss function as: $\mathcal{L}_{diff}=||\epsilon-\epsilon_\theta(x_t,t)||^2$, where $\theta$ represents the learnable parameters of diffusion network.

	\section{Methodology}
	In this section, we will introduce the detailed structure of DiffTF++. It consists of two stages: 1) triplane fitting~\ref{ff} and 2) diffusion training~\ref{zz}. The training pipeline of DiffTF++ is shown in Fig.~\ref{fig:conv}.

	\subsection{3D Representation Fitting}\label{ff}
	
	An efficient and robust 3D representation is of significance to train a 3D diffusion model for large-vocabulary object generation. In this paper, we adopt the triplane representation inspired by~\cite{chan2022efficient}. Specifically, it consists of three axis-aligned orthogonal feature planes, denoted as $F_{xy},F_{yz},F_{xz} \in \mathbb{R}^{C \times W \times H}$, and a multilayer perceptron (MLP) decoder for interpreting the sampled features from planes. By projecting the 3D coordinates on the triplane, we can query the corresponding features $F(p)=\mathrm{Cat}(F_{xy}(p),F_{yz}(p),F_{xz}(p))\in \mathbb{R}^{C \times 3}$ of any 3D position $p \in \mathbb{R}^{3}$ via bilinear interpolation. given the position $p \in \mathbb{R}^{3}$ and view direction $d \in \mathbb{R}^{2}$, the view-dependent color $c \in \mathbb{R}^{3}$ and density $\sigma \in \mathbb{R}^{1}$ can be formulated as: 
	\begin{equation}\label{e1}
		\begin{aligned}
			c(p,d),\sigma(p)=\mathrm{MLP}_{\theta}(F(p),\gamma(d))
		\end{aligned}
		~ ,
	\end{equation}
 where $\gamma (\cdot)$ represents positional embedding.
	 Detail structure of $\mathrm{MLP}_{\theta}$ is shown in Fig.~\ref{structure}.
	
	

	Since the fitted triplane is the input of diffusion, the triplanes of diverse real-world objects should be in the same space. Therefore, we need to train a robust category-independent shared decoder. Besides, to constrain the values of triplanes, we adopt strong L2 regularization and TVloss regularization~\cite{dey2006richardson} as follows: 
	\begin{equation}
		\begin{aligned}
			TV(F(p))=&\sum_{i,j}|F(p)[:,i+1,j]-F(p)[:,i,j]|+\\
   &|F(p)[:,i,j+1]-F(p)[:,i,j]|.
		\end{aligned}
	\end{equation}
	Additionally, to further accelerate the speed of triplane fitting, the $p$ merely samples from the points in the foreground. To avoid the low performance in predicting background RGB, we set a threshold to control the ratio of only sampling from the foreground or sampling from the whole image. Besides, we introduce a foreground-background mask. According to the volume rendering\cite{max1995optical}, the final rendered images $\hat{C}(p,d)$ and foreground mask $M(p)$ can be achieved by:
	$\hat{C}(p,d)=\sum_{i=1}^{N}T_i(1-\mathrm{exp}(-\sigma_i\delta_i))c_i, where~ T_i=\mathrm{exp}(-\sum_{j=1}^{i-1}\sigma_i\delta_i), \hat{M}(p)=\sum_{i=1}^{N}T_i(1-\mathrm{exp}(-\sigma_i\delta_i))$ Therefore, the training objective of triplane fitting can be formulated as:
	\begin{equation}\label{e3}
		\begin{aligned}
			\mathcal L = &\sum_{i}^{M} (\mathrm{MSE}(\hat{C}(p,d),{G}_c)+ \mathrm{MSE}(\hat{M}(p,d),{G}_{\sigma}))\\
   &+\lambda_1(\mathrm{TV}(F(p)))
			+\lambda_2(||F(p)||_2^2)
		\end{aligned}
		~ ,
	\end{equation}
	where ${G}_c$ and ${G}_\sigma$ represent the ground-truth of RGB and alpha while $\lambda_1$ and $\lambda_2$ are coefficients.
	


	\subsection{3D-aware Diffusion with transformer}\label{zz}

 
	\subsubsection{3D-aware encoder/decoder}
	
	In general, low-resolution features tend to contain richer semantic information and exhibit greater robustness, while high-resolution features are beneficial for capturing detailed features. Therefore, to enhance the 3D-related information in triplanes and avoid compute-intensive operation in the 3D-aware transformer, we propose a new 3D-aware encoder/decoder to encode the original triplanes $F^t=\mathrm{Cat}(F_{xy}^t,{F}_{yz}^t,{F}_{xz}^t) \in \mathbb{R}^{C \times W \times 3H}$ with 3D awareness. 
	
	To avoid compute-intensive operations in the encoder/decoder, we adopt a single convolution to decrease the resolution of features. Note that since the triplanes have 3D relations, the convolution should be performed on individual planes respectively. For simplification, we denote the output of convolution as $\overline{F}^t \in \mathbb{R}^{C \times W' \times 3H'}, W'=W/n, H'=H/n$, where $n$ represents the stride of convolution as shown in Fig~\ref{fig:details}. Following the patchify module in~\cite{dosovitskiy2020image}, we adopt a patchify layer (denote as $\mathcal{G}$) to convert the spatial features into a sequence of $M$ tokens. Thus, given the patch size $ps$, the dimension of output feature sequences can be formulated as: $\overline{F}' =\mathcal{G}(\overline{F}^t)\in \mathbb{R}^{C \times 3M}$, where $M=W'/ps*H'/ps$. Since the attention operation is position-independent, the learnable positional embedding is applied after the patchify module (denoted the output as $\overline{F}$). Considering triplanes from different categories may have differences in feature space, we introduce the conditional information using an adaptive norm layer inspired by~\cite{peebles2022scalable}.
	
	To extract the generalized 3D knowledge across different planes, we introduce the cross-plane attention whose details are shown in Fig.~\ref{fig:detail}. It takes the target plane (for example xy-plane) as the query while treating the whole triplane as key and values. The 3D constraint in feature space between xy-plane and three planes can be obtained as:
	\begin{equation}
		\begin{aligned}
			&\widetilde{F}_{xy}=\overline{F}_{xy}+\alpha_1*\mathrm{MultiHead}(\overline{F}_{xy},{\overline{F}},{\overline{F}})
		\end{aligned}
		~ ,
	\end{equation}
	where $\alpha_1$ represents the calibration vector obtained by a MLP. To maintain the 3D-related information, we adopt a share-weight attention module. Consequently, the encoded feature with 3D awareness can be concatenated: $\widetilde{F}=\mathrm{Cat}(\widetilde{F}_{xy},\widetilde{F}_{yz},\widetilde{F}_{xz})$, where $\widetilde{F}_{yz}$ and $\widetilde{F}_{xz}$ are the 3D constraint features from yz and xz planes respectively.
	
	In the end, the sequence features can be restored to spatial features via an integration layer (denoted as $\mathcal{G}^{-1}$): $\widetilde{F}^{t}=\mathcal{G}^{-1}(\widetilde{F}) \in \mathbb{R}^{C \times W' \times H'}$.

	\subsubsection{3D-aware Transformer}
	
	Benefiting from the 3D-aware transformer encoder, the 3D-related information is enhanced in the encoded high-level features. To extract the global generalized 3D knowledge and aggregate it with specialized features further, we build the 3D-aware transformer illustrated in Fig.~\ref{fig:detail}. It consists of self- and cross-plane attention. The former aims to enhance the specialized information of individual planes while the latter concentrates on building generalized 3D prior knowledge across different planes. 

	Similar to the 3D-aware encoder, we adopt the patchify module and the learnable positional embedding. For clarification, we denote the input of the 3D-aware transformer as $\hat{F}$. Since the encoded features have in terms of plane-dependent and plane-independent information, we build the transformer in two branches.
	Specifically, the self-attention module in the left branch can build interdependencies within individual planes. By exploiting 2D high-level features in different planes, our transformer will exploit the rich semantic information. Therefore, the output of enhanced features can be formulated as follows:
	\begin{equation}
		\begin{aligned}
			&\hat{F}^{s}=\mathrm{Norm}(\hat{F}_1+\mathrm{MLP}(\hat{F}_1))~,\\
			&\hat{F}_1=\mathrm{Norm}(\hat{F}+\mathrm{MultiHead}(\hat{F},\hat{F},\hat{F}))
		\end{aligned}
		~ ,
	\end{equation}
	where $\mathrm{Norm}$ represents the layer normalization.

	As for the right branch, it focuses on extracting the global 3D relations across different planes. Similarly, to enhance the explicit 2D features in each plane, we adopt an additional self-attention module before cross-plane attention. Meanwhile, to avoid the negative influence of 2D semantic features, we apply the residual connection. Consequently, the 3D-related information of xy planes can be formulated as:
	\begin{equation}
		\begin{aligned}
			&\hat{F}^{final}_{xy}=\mathrm{Norm}(\hat{F}^{c}_{xy}+\mathrm{MLP}(\hat{F}^{c}_{xy})\\
			&\hat{F}^{c}_{xy}=\mathrm{Norm}(\hat{F}^{*}_{xy}+\mathrm{MultiHead}(\hat{F}^{*}_{xy},\hat{F}^{s},\hat{F}^{s}))\\
			&\hat{F}^{*}_{xy}=\mathrm{Norm}(\hat{F}_{xy}+\mathrm{MultiHead}(\hat{F}_{xy},\hat{F}_{xy},\hat{F}_{xy}))
		\end{aligned}
		~ .
	\end{equation}

	By concatenating the features from three planes and integration layer ($\mathcal{G}^{-1}$) similar to 3D-aware encoder, the final features containing 3D-related information can be restored to the spatial features:$\hat{F}_{d}=\mathcal{G}^{-1}(\mathrm{Cat}(\hat{F}^{final}_{xy},\hat{F}^{final}_{yz},\hat{F}^{final}_{xz})) \in \mathbb{R}^{C \times W' \times H'}$.
	
\subsubsection{3D-aware Refinement and multi-view reconstruction loss}
In order to filter out the noisy information from triplanes robustly and efficiently, we adopt a similar architecture to our proposed 3D-aware encoder/decoder. Note that different from the objective of predicting noise in the diffusion model, the input of our refinement is generated triplane rather than noise. By utilizing interdependencies among different planes, the refinement can eliminate the outliers and artifacts efficiently. For clarification, we denote the input of the refinement as $F^{raw} \in \mathbb{R}^{C \times W \times H}$. To avoid decreasing performance, we adopt a residual connection. Thus, the refined triplane can be formulated as:
\begin{equation}
		\begin{aligned}
			&F^{ref}=F^{raw}+\mathcal{R}_\phi(F^{raw})
		\end{aligned}
		~ ,
	\end{equation}
where $\phi$ represents the parameters of 3D-aware refinement. 

Obtaining the refined triplane, we will sample $N_{render}$ points (denoted as $P_{r} \in \mathbb{R}^{N_{render} \times 3}$) randomly and query the corresponding features by projecting the 3D coordinates on the refined triplanes. Therefore, the multi-view reconstruction loss can be expressed as follows:

\begin{equation}
		\begin{aligned}
			&\mathcal{L}_{loss}= \mathrm{MSE}(c^{ref}(P_{r},d_r),G_c)\\
			&c^{ref}(P_{r},d_r),\sigma(P_{r})=\mathrm{MLP}_{\theta}(F^{ref}(P_{r}),\gamma(d_r))\\
			&F^{ref}(P_{r})=\mathrm{Cat}(F^{ref}_{xy}(P_{r}),F^{ref}_{yz}(P_{r}),F^{ref}_{xz}(P_{r}))
		\end{aligned}
		~ ,
	\end{equation}
where $d_r \in \mathbb{R}^{N_{render} \times 3}$ represents the view direction.

Therefore the total loss function for diffusion training can be obtained as:
\begin{equation}
		\begin{aligned}
			&\mathcal{L}_{diffimg}= \lambda_3\mathcal{L}_{diff}+\lambda_4\mathcal{L}_{loss}
		\end{aligned}
		~ ,
	\end{equation}
 where $\lambda_3$ and $\lambda_4$ are coefficients to balance the two branches. 
  
  
  
   

 \section{Experiments}
	\subsection{Implementation details}
 \subsubsection{Data processing} 
 
 Following most previous works, we use the ShapeNet~\cite{chang2015shapenet} including Chair, Airplane, and Car for evaluating the 3D generation which contains 6770, 4045, and 3514 objects, respectively. Additionally, to evaluate the large-vocabulary 3D object generation, we conduct the experiments on a most recent 3D dataset, OmniObject3D~\cite{wu2023omniobject3d}. OmniObject3D is a large-vocabulary real scanned 3D dataset, containing 216 challenging categories of 3D objects with high quality and complicated geometry, \textit{e.g.}, toy, fruit, vegetable, and art sculpture. To train the triplane and shared decoder on ShapeNet, we use the blender to render the multi-view images from 195 viewpoints. Those points sample from the surface of a ball with a 1.2 radius. Similarly, we use the blender to render the 5900+ objects from 100 different viewpoints to fit the triplane and decoder on OmniObject3D following~\cite{wu2023omniobject3d}.


		


 \begin{table}[b]
		\centering
		\caption{\textbf{Quantitative comparison of conditional generation on the OmniObject3D dataset.} DiffTF outperforms other SOTA methods in terms of 2D and 3D metrics by a large margin.}
		\renewcommand\tabcolsep{2pt}
		\small
		\begin{tabular}{lcccc}
			\toprule
			Methods & FID$\downarrow$&KID(\%)$\downarrow$&COV(\%)$\uparrow$&MMD(\textperthousand)$\downarrow$   \\
			\midrule
			EG3D~\cite{chan2022efficient}&41.56&1.0&14.14&28.21   \\
			GET3D~\cite{gao2022get3d}&49.41&1.5&24.42&13.57  \\
			DiffRF~\cite{muller2022diffrf}&147.59&8.8&17.86&16.07   \\
			NFD w/ texture~\cite{shue20223d}&122.40&4.2&33.15&10.92   \\
			
			\midrule
			\textbf{DiffTF (Ours)}  &\textbf{25.36}&\textbf{0.8}&\textbf{43.57}&\textbf{6.64}   \\
   \textbf{DiffTF++ (Ours)}  &\textbf{20.97}&\textbf{0.6}&\textbf{45.24}&\textbf{6.02}   \\

			\bottomrule
		\end{tabular}
		
		\label{tab:omni}%
	\end{table}%

\begin{figure*}[p]
	\centering

	\includegraphics[width=1\textwidth]{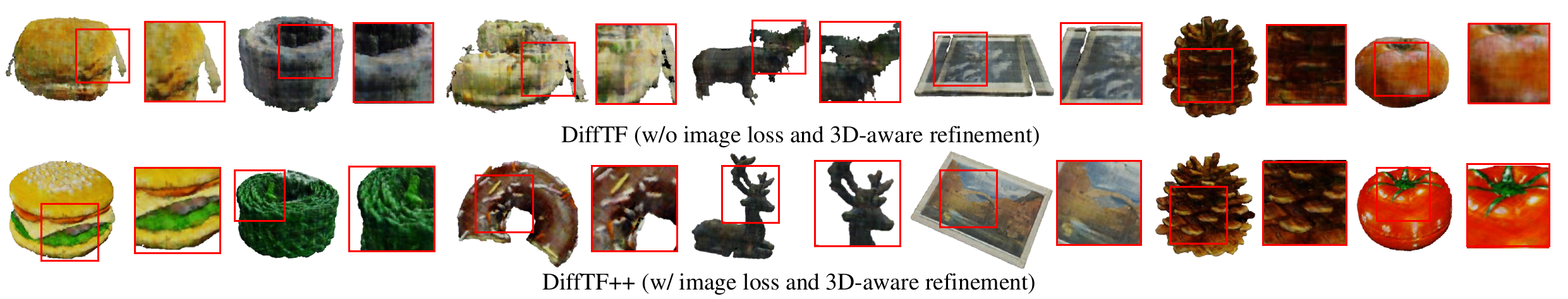}
	\vspace{-20pt}
	\caption{Detailed comparison between the DiffTF (without multi-view reconstruction loss function and 3D-aware refinement) and DiffTF++. It proves that our proposed modules can filter out the noise information, thereby eliminating the artifacts in generated 3D objects effectively. Furthermore, our refinement can enrich the details of the generated topology and generate high-quality 3D objects with abundant texture.}\label{fig:detailomni2}
	
\vspace{10pt}
 \includegraphics[width=1\textwidth]{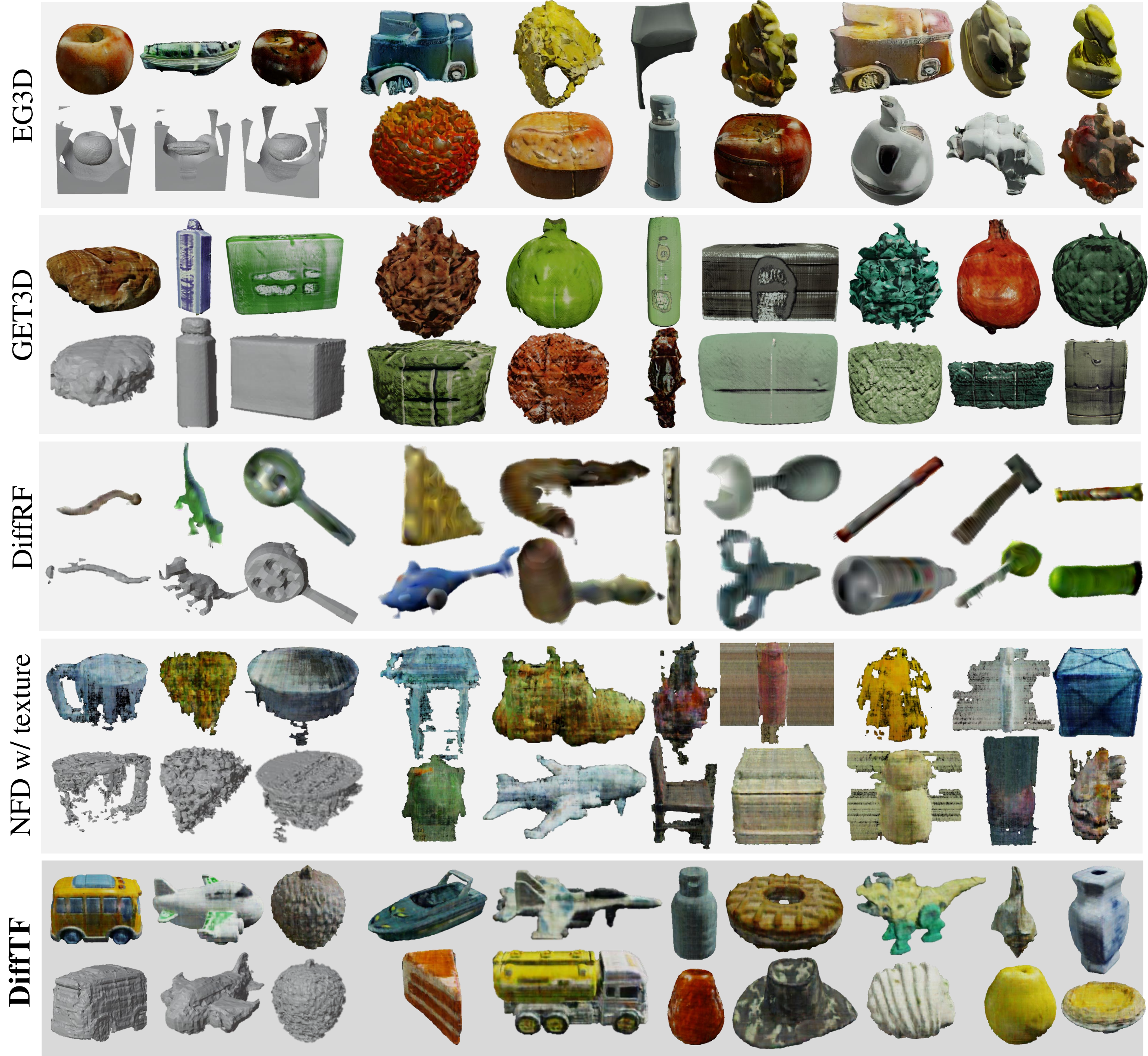}
		
		\caption{\textbf{Qualitative comparisons to the SOTA methods in terms of generated 2D images and 3D shapes on OmniObject3D.} Compared with other SOTA methods, our generated results are more realistic with richer semantics. }\label{fig:com}

\end{figure*}
  
\begin{table*}[t]
		\centering
		
		\caption{\textbf{Quantitative comparison of unconditional generation on the ShapeNet}. It proves the impressive performance of our DiffTF on the single-category generation.}
		\small
        \renewcommand{\arraystretch}{1.2}
		\begin{tabular}{lccccc}
			\toprule
			Category&Method&FID$\downarrow$&KID(\%)$\downarrow$&COV(\%)$\uparrow$&MMD(\textperthousand)$\downarrow$   \\
			\midrule
			\multirow{6}*{Car}&EG3D~\cite{chan2022efficient}& 45.26&2.2&35.32& 3.95  \\
			&GET3D~\cite{gao2022get3d} & 41.41&1.8&37.78&3.87  \\
			
			&DiffRF~\cite{muller2022diffrf}& 75.09&5.1&29.01&4.52   \\
			&NFD w/ texture~\cite{shue20223d}& 106.06&5.5&39.21&3.85  \\
   
			\cline{2-6} 
			
			&\textbf{DiffTF (Ours)} & \textbf{36.68}&\textbf{1.6}&\textbf{53.25}& \textbf{2.57}  \\

   &\textbf{DiffTF++ (Ours)} & \textbf{32.74}&\textbf{1.4}&\textbf{54.41}& \textbf{2.34}  \\

			\midrule
			\multirow{6}*{Plane}&EG3D~\cite{chan2022efficient}& 29.28&1.6&18.12&4.50   \\
			&GET3D~\cite{gao2022get3d} & 26.80&1.7&21.30&4.06  \\
			&DiffRF~\cite{muller2022diffrf}& 101.79&6.5&37.57&3.99   \\
			&NFD w/ texture~\cite{shue20223d}& 126.61&6.3&34.06&2.92   \\

   \cline{2-6} 
			&\textbf{DiffTF (Ours)} & \textbf{14.46}&\textbf{0.8}&\textbf{45.68}&\textbf{2.58}  \\

   &\textbf{DiffTF++ (Ours)} & \textbf{11.32}&\textbf{0.8}&\textbf{46.32}&\textbf{2.23}  \\

			\midrule
			
			\multirow{6}*{Chair}&EG3D~\cite{chan2022efficient}& 37.60&2.0&19.17&10.31   \\
			&GET3D~\cite{gao2022get3d} & 35.33&1.5&28.07& 9.10  \\
			&DiffRF~\cite{muller2022diffrf}& 99.37&4.9&17.05&14.97   \\
			&NFD w/ texture~\cite{shue20223d}& 87.35&2.9&31.98&7.12   \\
   
			\cline{2-6} 

			&\textbf{DiffTF (Ours)} & \textbf{35.16}&\textbf{1.1}&\textbf{39.42}&\textbf{5.97}  \\

   &\textbf{DiffTF++ (Ours)} & \textbf{31.52}&\textbf{1.0}&\textbf{40.23}&\textbf{5.46}  \\
 
			\bottomrule
		\end{tabular}

		\label{tab:car}%
	\end{table*}%
	\begin{figure}[t]
	\centering

	\includegraphics[width=0.5\textwidth]{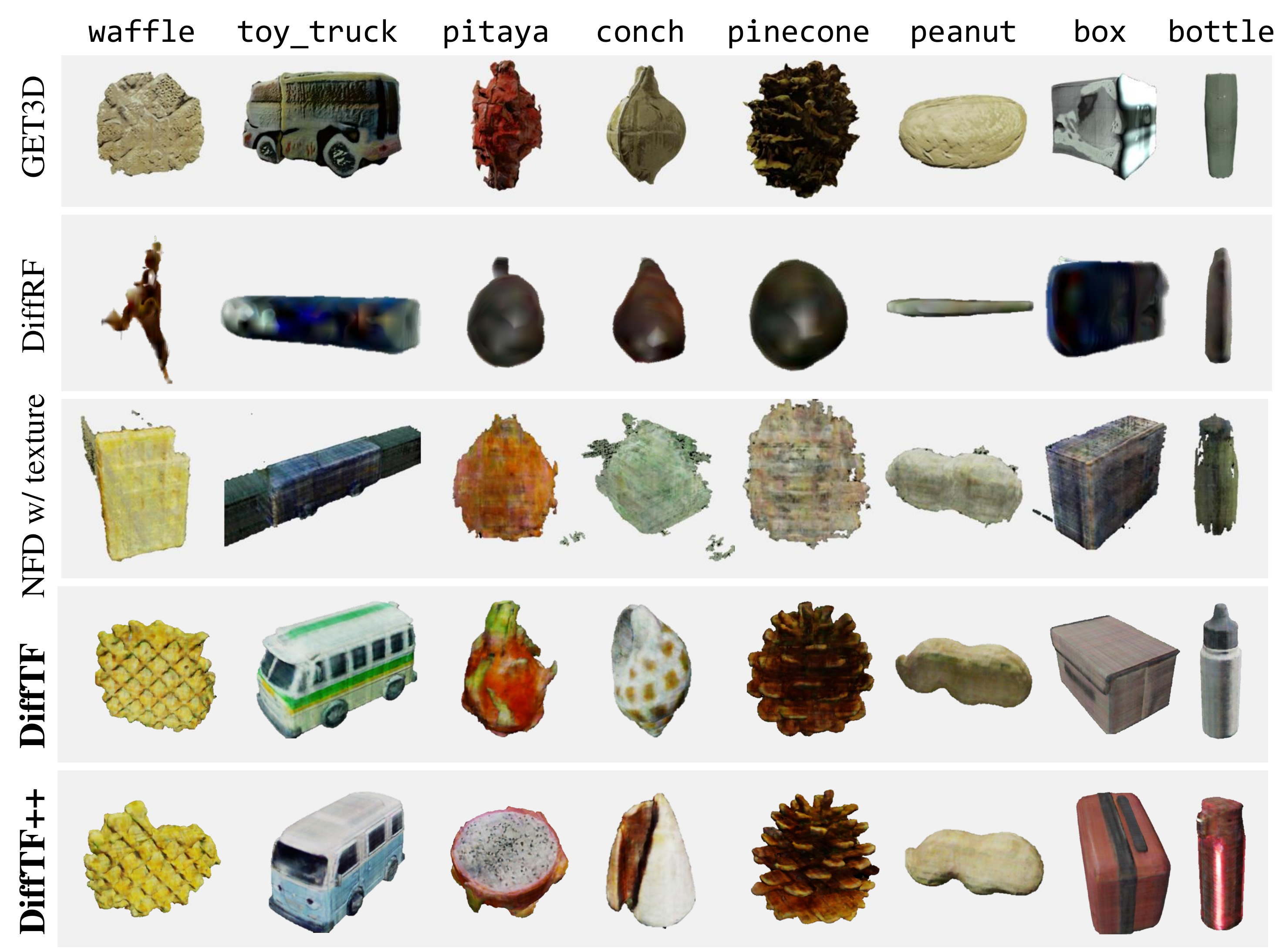}
	
	\caption{Comparison against other methods on class-conditional generation.}\label{fig:com_cate}
	
	\vspace{-10pt}
	
\end{figure}
\begin{figure}[t]
		\centering	
		
		\includegraphics[width=0.5\textwidth]{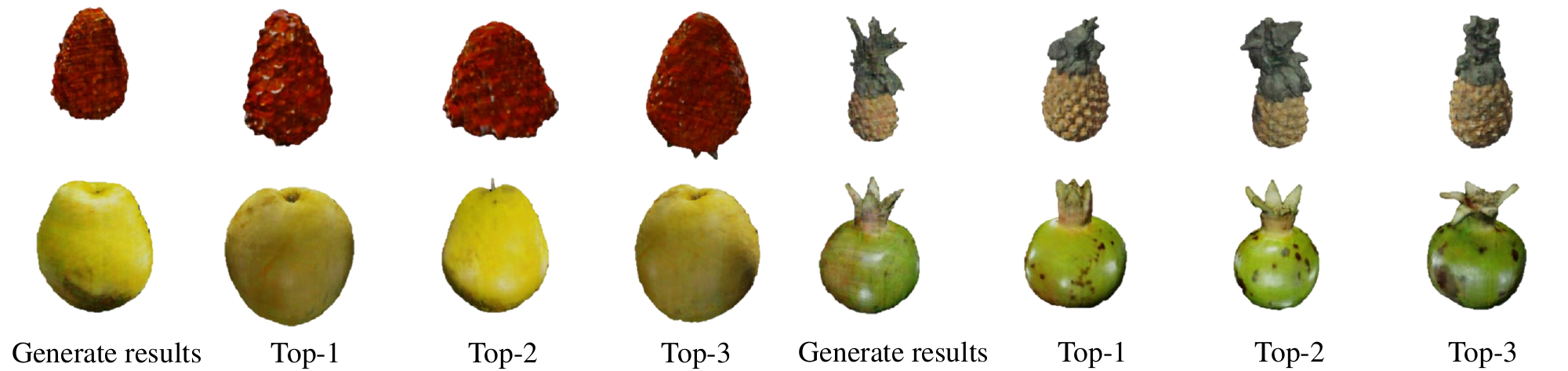}
		\vspace{-10pt}
		\caption{Nearest Neighbor Check on OmniObject3D. We compare our generated results and the most similar top 3 objects from the training set.}\label{nnc}

  \includegraphics[width=0.5\textwidth]{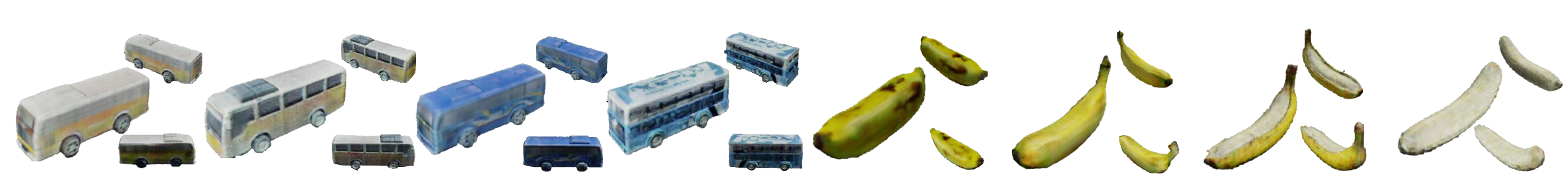}
		
		\caption{Interpolation between generated results on OmniObject3D.}\label{inte}

	\end{figure}
\begin{figure*}[p]
	\centering

	\includegraphics[width=1\textwidth]{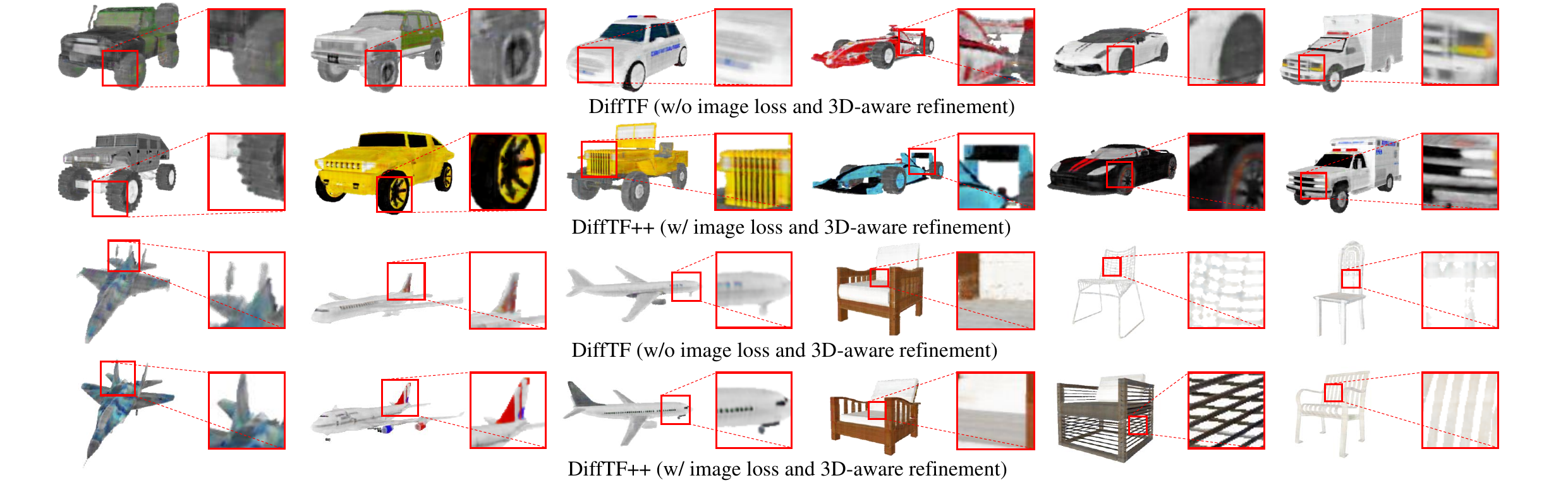}
	\vspace{-20pt}
	\caption{Detailed comparison between the DiffTF (without multi-view reconstruction loss function and 3D-aware refinement) and DiffTF++ on ShapeNet. It clearly shows that our refinement and multi-view reconstruction loss can improve the overall generative performance, especially in detailed topology and textures.}
	\label{com_shapenet1}
	\vspace{10pt}

 \includegraphics[width=1\textwidth]{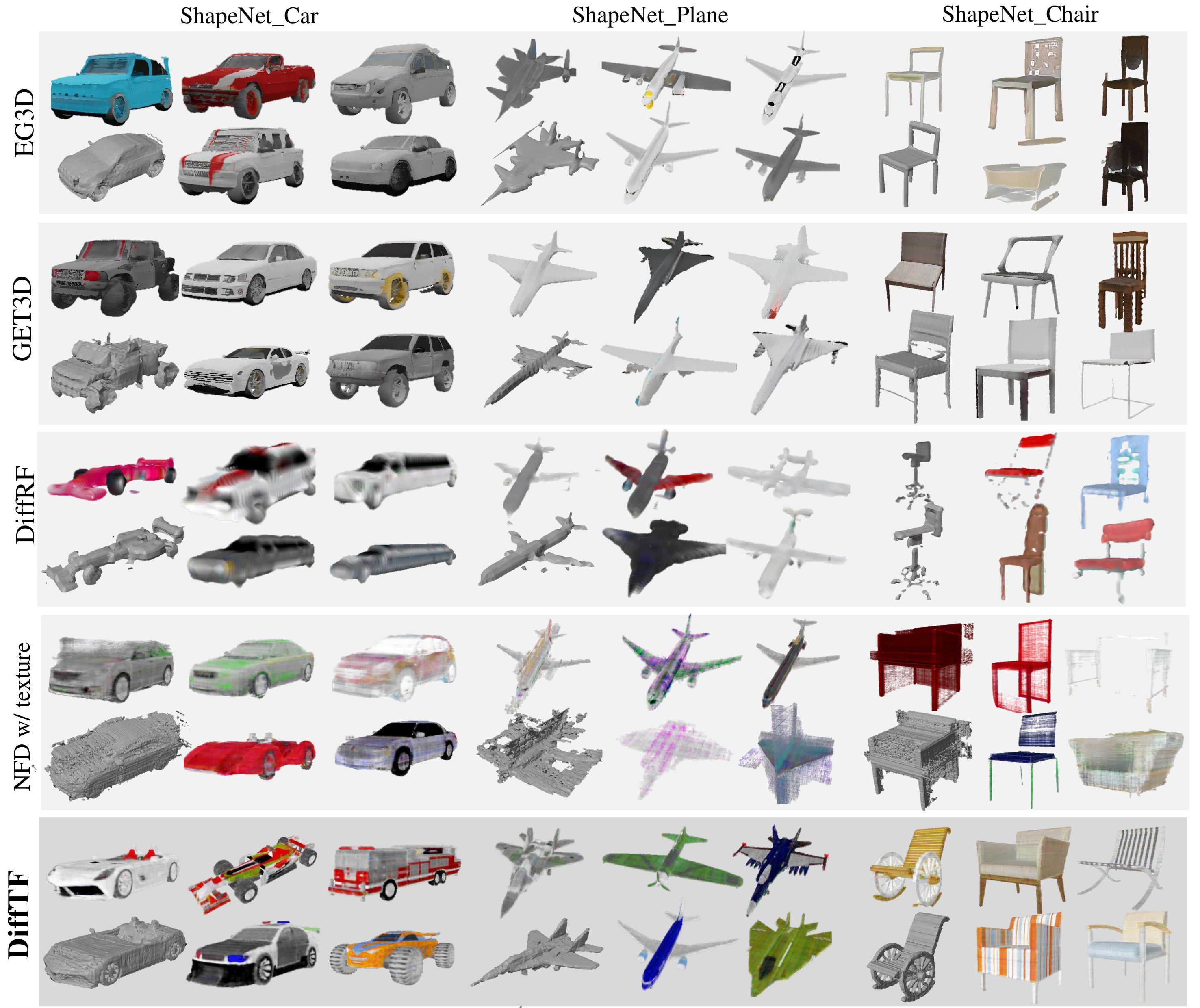}
		\caption{\textbf{Qualitative comparison of DiffTF against other SOTA methods on ShapeNet}. It intuitively illustrates the promising performance of our method in texture and topology. }
		\label{fig:com_omni}
	
\end{figure*}

 \subsubsection{Two-stage Training}
	The training process consists of two steps: \textbf{I}) triplane fitting and \textbf{II}) diffusion traning. In the first step, \textit{i.e.}, triplane fitting, the objective is to obtain the diverse triplane features and robust triplane decoder. Therefore, for clarification, we divide \textbf{I} into two subtasks: \textbf{step I-I} training shared decoder and \textbf{step I-II} optimizing triplanes from diverse 3D objects. To maintain the robustness of the decoder, we adopt around 20 percent diverse and high-quality objects for optimizing the shared decoder in the \textbf{step I-I}. Then, in \textbf{step I-II}, we adopt the trained decoder with frozen parameters to merely fit the triplanes. After obtaining the fitted triplanes, we can use them as the ground truth to train the 3D-aware transformer-based diffusion model (\textbf{step II}).

 \textbf{Triplane fitting.} Our implementation is based on the PyTorch framework. The dimension of the triplane is $18 \times 256 \times 256$. Note that $\lambda_1$ and $\lambda_2$ are set to 1e-4 and 5e-5 for training the share-weight decoder. We train our shared decoder using 8 GPUs for 24 hours. After getting the decoder, $\lambda_1$ and $\lambda_2$ are set to 0.5 and 0.1 for triplane fitting. To improve the robustness of the shared decoder, we adopt the one-tenth learning rate (1e-2) during the training while the learning rate of the triplane feature is set to 1e-1.

 \textbf{Diffusion and refinement.} We adopt the cross-plane attention layer in the 3D-aware encoder when the feature resolution is 64, 32, and 16. We adopt 8, 4, and 2 as the patch size in the encoder/decoder. The patch size and number of the 3D-aware transformer layers are set to 2 and 4, respectively. Following the prior work~\cite{muller2022diffrf,shue20223d}, we adopt T=1000 during training and T=250 for inference. Our diffusion model is trained using an Adam optimizer with a learning rate of 1e-4 which will decrease from 1e-4 to 1e-5 in linear space. We apply a linear beta scheduling from 0.0001 to 0.01 at 1000 timesteps. Besides, we adopt the $\epsilon$ as the objective of our diffusion model. The batch size of the diffusion model and number of sampling $N_{render}$ are set to 2 and 10000, respectively. As for the refinement module, we adopt 4 as the patch size of the refinement and 32 as the target resolution of the cross-plane attention layer. Note that before the first 300000 iteration, we froze the parameters of refinement and only updated the diffusion model. Meantime, $\lambda_4$ is set to 0 to avoid abrupt fluctuation. After that, we set $\lambda_3=-0$ while $\lambda_4=1$ to fine-tune the whole model. We fine-tune our model for about 2 days on 16 NVIDIA A100 GPUs. Codes are available at ~\url{https://github.com/ziangcao0312/DiffTF}.




   \subsubsection{Sampling procesing}
 Similar to the training process, sampling the 3D content from DiffTF has two individual steps: 1) using a trained diffusion model to denoise latent noise into triplane features, and 2) adopting the trained triplane decoder to decode the implicit features into the final 3D content.
 
 \textbf{Details about Interpolation} \cite{song2020denoising} proves smooth interpolation
	in the latent space of diffusion models can be achieved by
	interpolation between noise tensors before they are iteratively
	denoised by the model. Therefore, we sample from
	our model using the DDIM method. To guarantee the same distribution of the interpolation samples, we adopt spherical interpolation.

	\subsubsection{Evaluation Metrics}
 
 Following prior work~\cite{muller2022diffrf,shue20223d}, we adopt two well-known 2D metrics and two 3D metrics: a) Fr$\acute{\mathrm{e}}$chet
	Inception Distance~\cite{heusel2017gans} (FID-50k) and Kernel Inception Distance~\cite{binkowski2018demystifying} (KID-50k); b) Coverage Score (COV) and Minimum Matching Distance (MMD) using Chamfer Distance (CD). All metrics are evaluated at a resolution of 128 $\times$ 128. More details about metric and other SOTA methods are released in Appendix~\ref{evaldetail}.

	\subsection{Comparison against state-of-the-art methods}
	In this section, we compare our methods with state-of-the-art methods, including two GAN-based methods: EG3D~\cite{chan2022efficient}, GET3D~\cite{gao2022get3d} and three diffusion-based methods: DiffRF~\cite{muller2022diffrf}, NFD~\cite{shue20223d}, and our previous method, \textit{i.e.}, DiffTF~\cite{cao2023large}.

	\subsubsection{Large-vocabulary 3D generation on OmniObject3D}

 In this subsection, we compare our new method against other SOTA methods. The class-conditional quantitative results on OmniObject3D are shown in Table~\ref{tab:omni}. It illustrates that compared with NFD w/texture that uses the 2D CNN diffusion, our 3D-aware transformer achieves promising improvement in both texture and geometry metrics, especially in texture metric. Also, it proves the effectiveness of the global 3D awareness introduced by our transformer. Compared to DiffRF adopting 3D CNN on voxel, the diffusion on triplane features can benefit from higher-resolution representation to boost performance. Our methods achieve an impressive improvement against the SOTA GAN-based methods. Additionally, plus 3D-aware refinement and imaged-based loss, DiffTF++ outperforms our previous version with an over 17\% improvement in FID. 
 
 Besides, we visualize the generated results to intuitively evaluate our method. On the one hand, relying on the strong ability of our proposed 3D-aware transformer, our DiffTF can extract robust generalized 3D knowledge for generation. Therefore, as shown in illustrated in Fig~\ref{fig:com}, our generated objects have reasonable topology and texture that make these 3D content realistic. On the other hand, we show the improvement brought by 3D-aware refinement and multi-view reconstruction loss shown in Fig~\ref{fig:vis}, and Fig~\ref{fig:detailomni2}. Our proposed modules can not only eliminate the small artifacts in the 3D assets, \textit{e.g.}, strawberry, toy\_plane, lemon, donut, and nipple, but complete the defect of the 3D assets, \textit{e.g.}, house, painting, and starfish. Furthermore, our proposed strategies enrich the detailed topology and improve the texture of generated 3D objects significantly, especially in tomato, doll, and toy\_animal. We also compared all class-conditional methods in OmniObject3D. As shown in Fig~\ref{fig:com_cate}, the generated 3D objects of DiffTF++ have more vivid texture and color.
 
 To validate the generative capability of our method, we perform the nearest neighbor check on OmniObject3D. As shown in Fig.~\ref{nnc}, our method can generate some novel objects. We achieve the nearest neighbor check via the CLIP model. After obtaining the CLIP features, we chose the top 3 results by measuring cosine distances. Additionally, Fig.~\ref{inte} shows that our latent space is relatively smooth.

	\begin{table*}[t]
		\centering
  \caption{\textbf{Ablation studies on OmniObject3D}. The \textit{TP Regu}, \textit{TP Norm}, \textit{Refinement}, and \textit{Reconstruction loss} represent the triplane regularization, triplane normalization, 3D-aware refinement, and multi-view reconstruction loss, respectively. }
		\renewcommand\tabcolsep{4pt}
		\small
		\begin{tabular}{ccccccccccc}
			\toprule
			TP Regu&TP Norm&CP& Ori-TF& CP-TF &Refinement&Reconstruction loss&FID$\downarrow$&KID(\%)$\downarrow$&COV(\%)$\uparrow$&MMD(\textperthousand)$\downarrow$   \\
			\midrule
			\XSolidBrush&\XSolidBrush&\XSolidBrush&\XSolidBrush&\XSolidBrush&\XSolidBrush&\XSolidBrush&125.21&2.6&21.31&16.10   \\
			\Checkmark&\XSolidBrush&\XSolidBrush&\XSolidBrush&\XSolidBrush&\XSolidBrush&\XSolidBrush&113.01&2.3&30.61&12.04  \\
			
			\Checkmark&\Checkmark&\XSolidBrush&\XSolidBrush&\XSolidBrush&\XSolidBrush&\XSolidBrush&109.81&2.0&33.15&10.92    \\
			\midrule
			\Checkmark&\Checkmark&\Checkmark&\XSolidBrush&\XSolidBrush&\XSolidBrush&\XSolidBrush&39.13&1.4&37.15&9.93   \\
			\Checkmark&\Checkmark&\Checkmark&\Checkmark  &N/A&\XSolidBrush&\XSolidBrush&52.35 &2.0&36.93&9.97  \\
			
			\Checkmark&\Checkmark&\Checkmark&N/A &\Checkmark &\XSolidBrush&\XSolidBrush&\textbf{25.36}&\textbf{0.8}&\textbf{43.57}&\textbf{6.64}   \\
			\Checkmark&\Checkmark&\Checkmark&N/A &\Checkmark &\XSolidBrush&\Checkmark&\textbf{23.30}&\textbf{0.8}&\textbf{43.67}&\textbf{6.51}   \\
			\Checkmark&\Checkmark&\Checkmark&N/A &\Checkmark &\Checkmark&\Checkmark&\textbf{20.97}&\textbf{0.6}&\textbf{45.24}&\textbf{6.02}   \\
			\bottomrule
		\end{tabular}

		\label{tab:deeper_avg}%
	\end{table*}%

\begin{figure*}[t]
		\centering

		\includegraphics[width=1\textwidth]{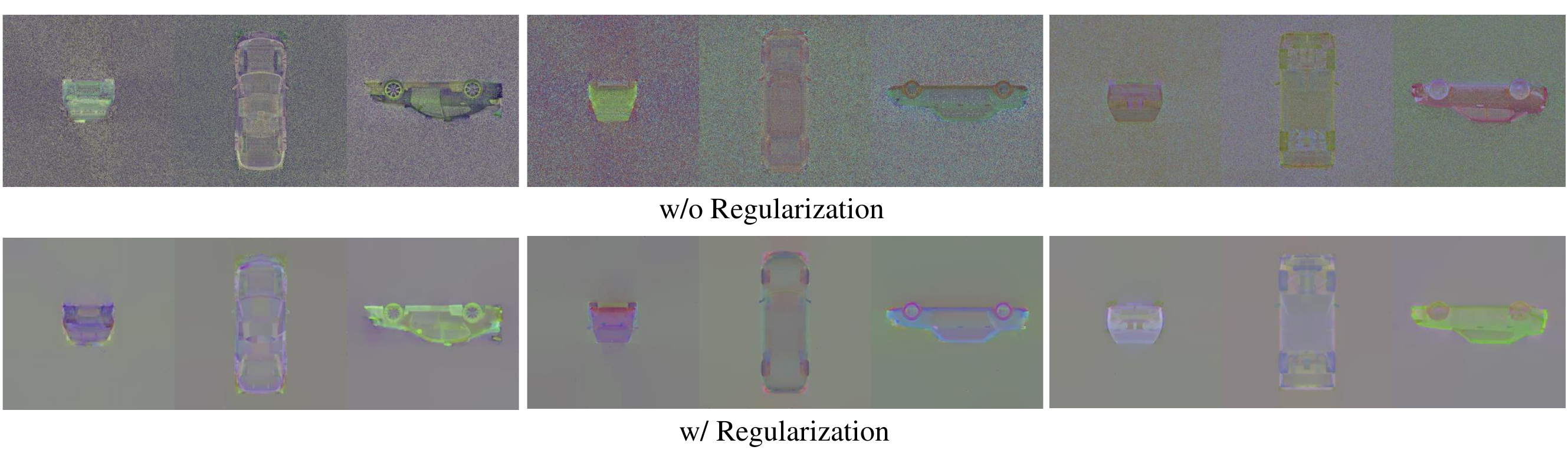}
		
		\caption{Visualization of triplane. Top: triplane fitting without TVloss and L2 regularization. Bottom: triplane fitting with TVloss and L2 regularization. It illustrates that by effective regularization, the triplane features are smooth and clear which is helpful for the next training.}
		\label{fig:addablation}
		\vspace{-10pt}
		
	\end{figure*}
 
	\begin{figure}[t]
		\centering

		\includegraphics[width=0.5\textwidth]{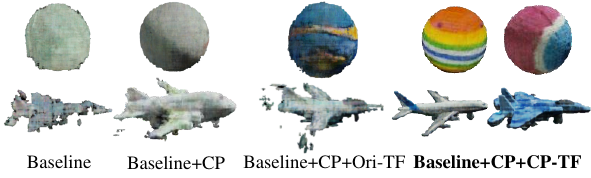}
  
		\caption{Ablations on 3D-aware modules. Note \texttt{Baseline+CP+CP-TF} represents our original method, \textit{i.e.}, DiffTF.}\label{abvis}
  \vspace{10pt}

		\includegraphics[width=0.5\textwidth]{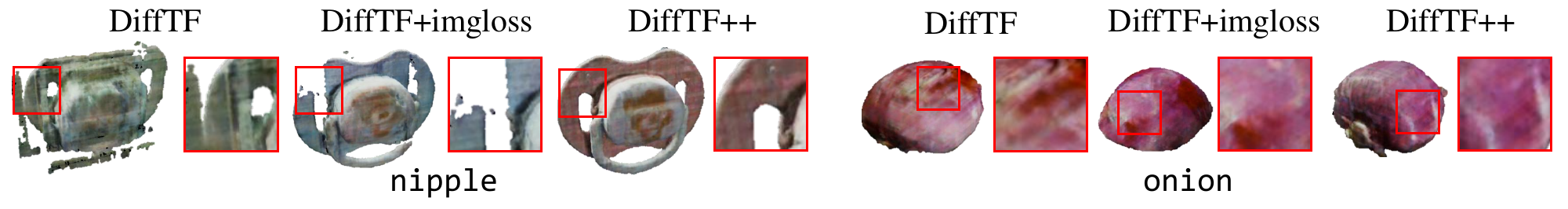}
  \vspace{-10pt}
		\caption{Ablations on 3D-aware refinement and multi-view reconstruction loss function. It shows that our proposed modules can eliminate the artifacts and improve the quality of the generated 3D objects in terms of topology and texture effectively.}\label{abvis1}
		\vspace{-20pt}
	\end{figure}

	\subsubsection{Single-category generation on ShapeNet}
	To further validate the performance of our methods, we evaluate them on ShapeNet~\cite{chang2015shapenet} in three categories. Table~\ref{tab:car} reports the quantitative comparison against other SOTA methods. It is clear that our methods achieve significant improvements in terms of all metrics in three categories. Similarly, we provide the visualization on ShapeNet~\cite{chang2015shapenet} in Fig~\ref{fig:com_omni} and Fig~\ref{com_shapenet1}. Depending on effective 3D awareness, our method DiffTF can generate 3D objects with rich semantical information, thereby standing out from all SOTA methods. However, generating reasonable details is still challenging. To handle this problem, we introduce the 3D-aware refinement and a multi-view reconstruction loss. As shown in Fig.~\ref{com_shapenet1}, they improve the detailed structure of the generated 3D objects significantly. DiffTF++ can generate more detailed structures, \textit{e.g.}, the special tread pattern, wheel hub, tail fins, and tiny windows. Also, it can eliminate the small artifacts in 3D objects (shown in line 3, column 4) and enrich the texture of small structures (illustrated in line 3, column 2). Furthermore, it can also maintain an impressive performance when generating tiny weaving structures (line 4, columns 5 and 6).

 \begin{figure*}[p]
		\centering
		\includegraphics[width=1\textwidth]{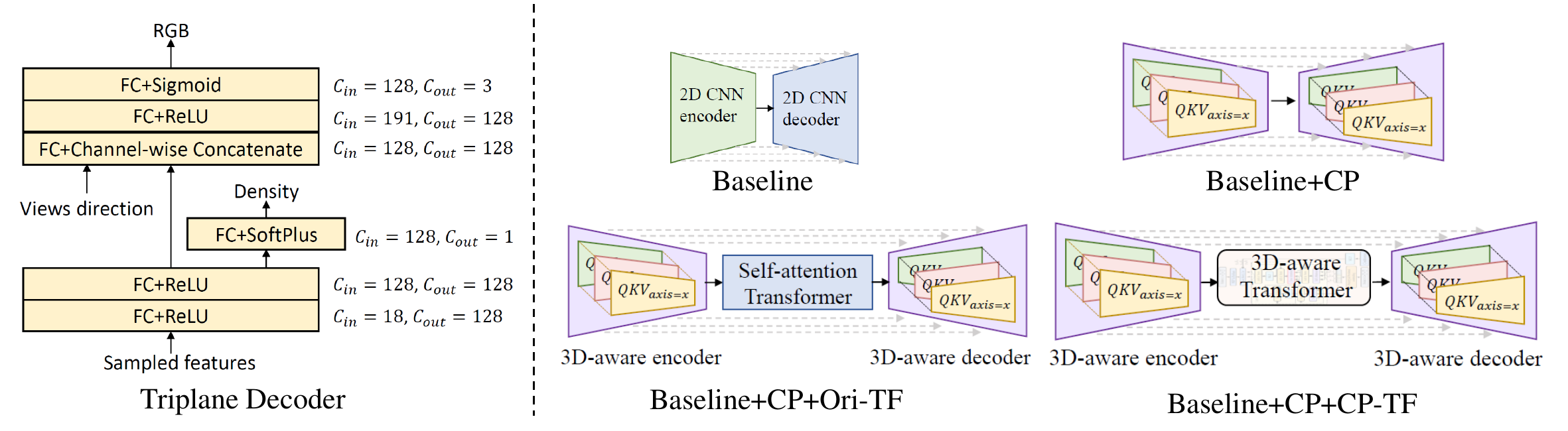}
  \caption{Details of our network used in experiments. Left: architecture of triplane decoder. Right: different structures of our method in ablation studies. }\label{structure}
  \includegraphics[width=0.8\textwidth]{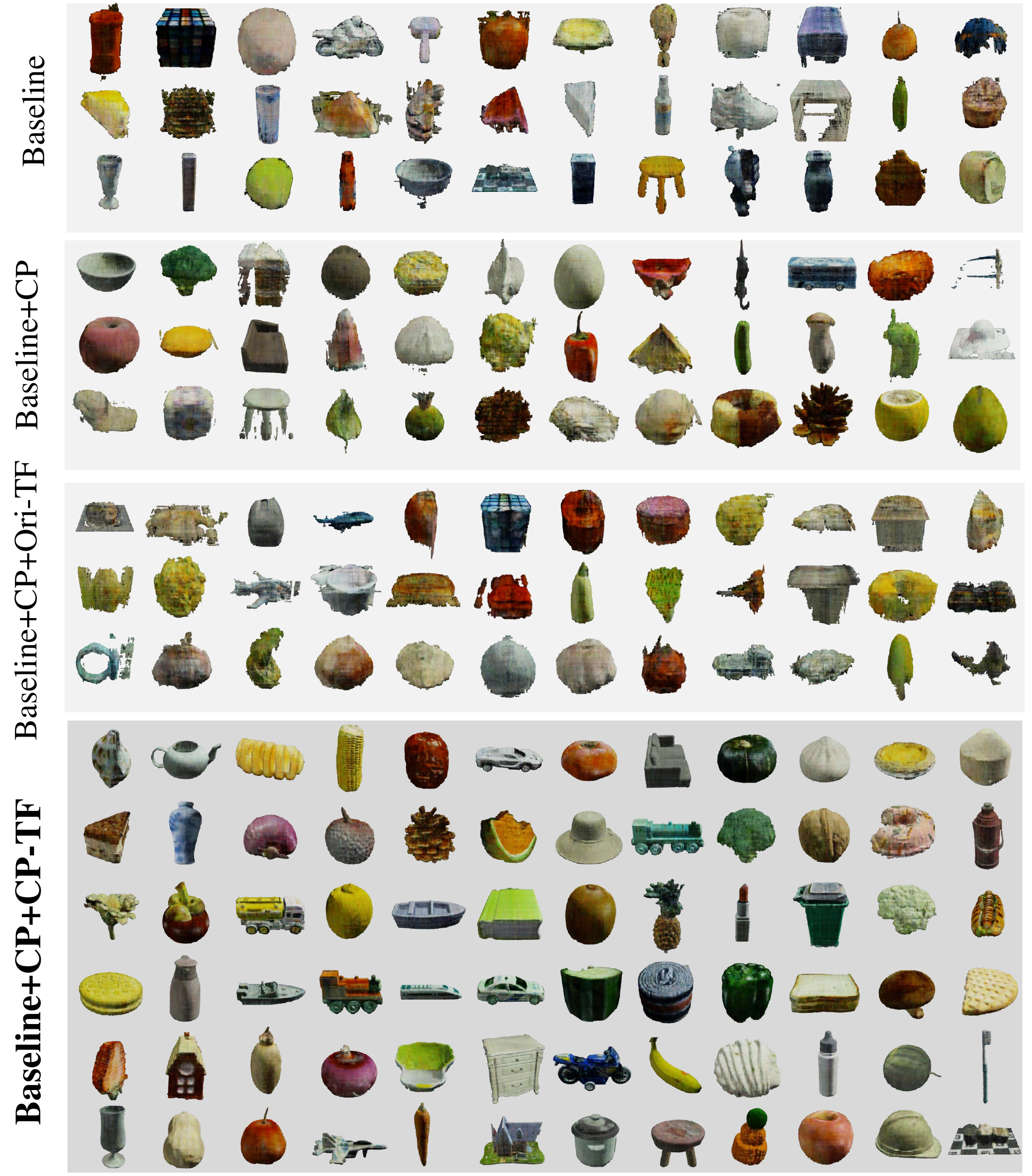}
  
		\caption{Qualitative comparison with different structures. Note \texttt{Baseline+CP+CP-TF} represents our original method, \textit{i.e.}, DiffTF.  }\label{abvis2}
		
	\end{figure*}

	\subsection{Ablation study}

	In this section, we conduct exhaustive ablation studies in several ways: 1) triplane fitting: normalization, regularization, and sampling strategy; 2) diffusion: our proposed 3D-aware modules and other 3D-aware strategies; and 3) 3D-aware refinement and loss function

	\subsubsection{Ablations about triplane fitting} 
 
\begin{figure}[t]
	\centering
	\vspace{-10pt}
		\includegraphics[width=0.4\textwidth]{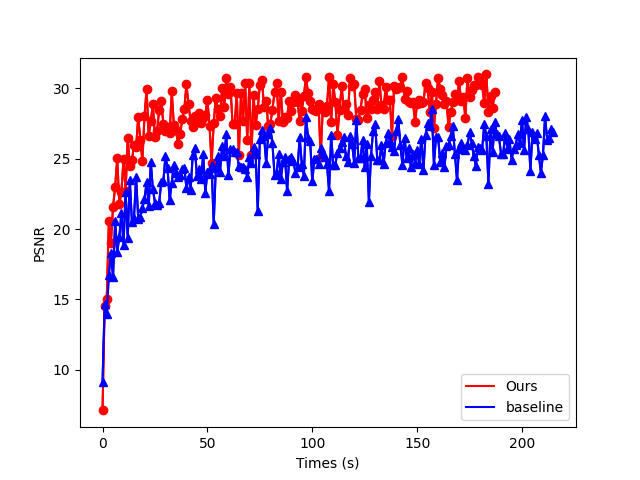}
		\caption{Ablations on sampling strategy. Our simple sampling strategy can accelerate the convergence effectively.}\label{tri}
	
\end{figure}

 Since the triplane fitting is the foundation of the diffusion model, we first study the influence of normalization and regularization in triplane fitting and diffusion training. As shown in Fig.~\ref{fig:addablation}, the triplane features are more smooth and clear. With effective regularization, the generated objects have better shapes and geometry information. Since the distribution of triplanes is overwide (from -10 to 10) without constraint, it is essential to adopt normalization to accelerate the convergence. As illustrated in Table~\ref{tab:deeper_avg}, by adopting the preprocess on triplane features, the diffusion model can get better generative performance. In addition, we study the effectiveness of the new sampling strategy in triplane fitting. Notably, the PSNR in Fig.~\ref{tri} is measured merely in the foreground area. With the new sampling strategy, the speed of our triplane fitting is raised further.

		\begin{table}[t]
		\centering
  \caption{Comparison of different 3D-aware modules.}
		\renewcommand\tabcolsep{4pt}
		\begin{tabular}{lcccc}
			\toprule
			Methods&FID$\downarrow$&KID(\%)$\downarrow$&COV(\%)$\uparrow$&MMD(\textperthousand)$\downarrow$   \\
			\midrule
			
			3D-aware Convolution~\cite{wang2022rodin} &84.55&4.5&33.79&11.47 \\
			\textbf{DiffTF (Ours)} &\textbf{25.36}&\textbf{0.8}&\textbf{43.57}&\textbf{6.64}\\

    \textbf{DiffTF++ (Ours)}  &\textbf{20.97}&\textbf{0.6}&\textbf{45.24}&\textbf{6.02}   \\

			\bottomrule
		\end{tabular}
		\vspace{-10pt}
		\label{tab:comrodin}%
	\end{table}%
	\subsubsection{Studies of 3D-aware diffusion model} 
 For clarification, we denote the proposed cross-plane attention in the encoder, original transformer and our 3D-aware transformer as \texttt{CP}, \texttt{Ori-TF}, and \texttt{CP-TF}. The detailed structures are shown in Fig.~\ref{structure}
 
 Compared with 2D CNN, our 3D-aware encoder/decoder can encode the triplanes while enhancing the 3D relations. As shown in Fig.~\ref{abvis}, attributing to the enhanced 3D features, it can raise the overall generative performance.  Notably, since the original transformer merely utilizes 2D self-attention, the introduction of the 2D interdependencies will break the latent 3D relations built by the encoder, thereby impeding the performance. In contrast to the original transformer, our 3D-aware transformer can effectively extract generalized 3D-related information across planes and aggregate it with specialized one for large-vocabulary generation. Therefore, our DiffTF can achieve better performance than \texttt{Baseline+CP+Ori-TF}. Besides class-conditional comparison, we report more qualitative results in Fig.~\ref{abvis2}. It validates intuitively the improvement brought by our 3D-aware modules.

 Additionally, we report the quantitative evaluation in Table~\ref{tab:deeper_avg}. By introducing \textit{CP}, the diffusion model has achieved a significant improvement (over \textbf{64\%} in FID). The introduction of \textit{CP-TF} boosts the generative performance further. In conclusion, attributed to extracted generalized 3D knowledge and specialized 3D features, our 3D-aware transformer is capable of strong adaptivity for large-vocabulary 3D objects. 
 
 Furthermore, we also compare our methods with other 3D-aware modules, \textit{i.e.}, 3D-aware convolution~\cite{wang2022rodin}. Experiment results in Table.~\ref{tab:comrodin} demonstrate the impressive generative performance of our novel 3D-aware modules compared with 3D-aware convolution in large-vocabulary 3D generation.

 \subsubsection{Ablations about 3D-aware refinement and loss function} 
 
 To study the effectiveness of our 3D-aware refinement and the multi-view reconstruction loss, we report the metrics in Table.~\ref{tab:comrodin}. By adding the new loss and refinement, our DiffTF++ achieves increasing performance in terms of 2D and 3D metrics. We also release the qualitative comparison in Fig.~\ref{abvis1}. As shown in this figure, since the multi-view reconstruction loss belongs to 2D supervision, it can improve the texture but is not very effective in promoting topology. By introducing 3D-aware refinement, our DiffTF++ can generate some complex structures with reasonable and impressive texture.

		\begin{table}[t]
		\centering
  \caption{User study of DiffTF++ and other SOTA methods. The rating is on a scale of 0-100, the higher the better.}
		\begin{tabular}{lc}
			\toprule
			Methods&Overall Score (\%)$\uparrow$\\
			\midrule
			EG3D~\cite{chan2022efficient}&10.04\\
			GET3D~\cite{gao2022get3d}&10.19\\
               DiffRF~\cite{muller2022diffrf}& 6.80\\
			NFD w/ texture~\cite{shue20223d} &13.71\\
   \midrule
			\textbf{DiffTF (Ours)~\cite{cao2023large}} &18.97\\
   \textbf{DiffTF++ (Ours)} &40.28\\

			\bottomrule
		\end{tabular}
		\vspace{-10pt}
		\label{tab:user}%
	\end{table}%
	\begin{figure}[t]
	\centering
	
		\includegraphics[width=0.5\textwidth]{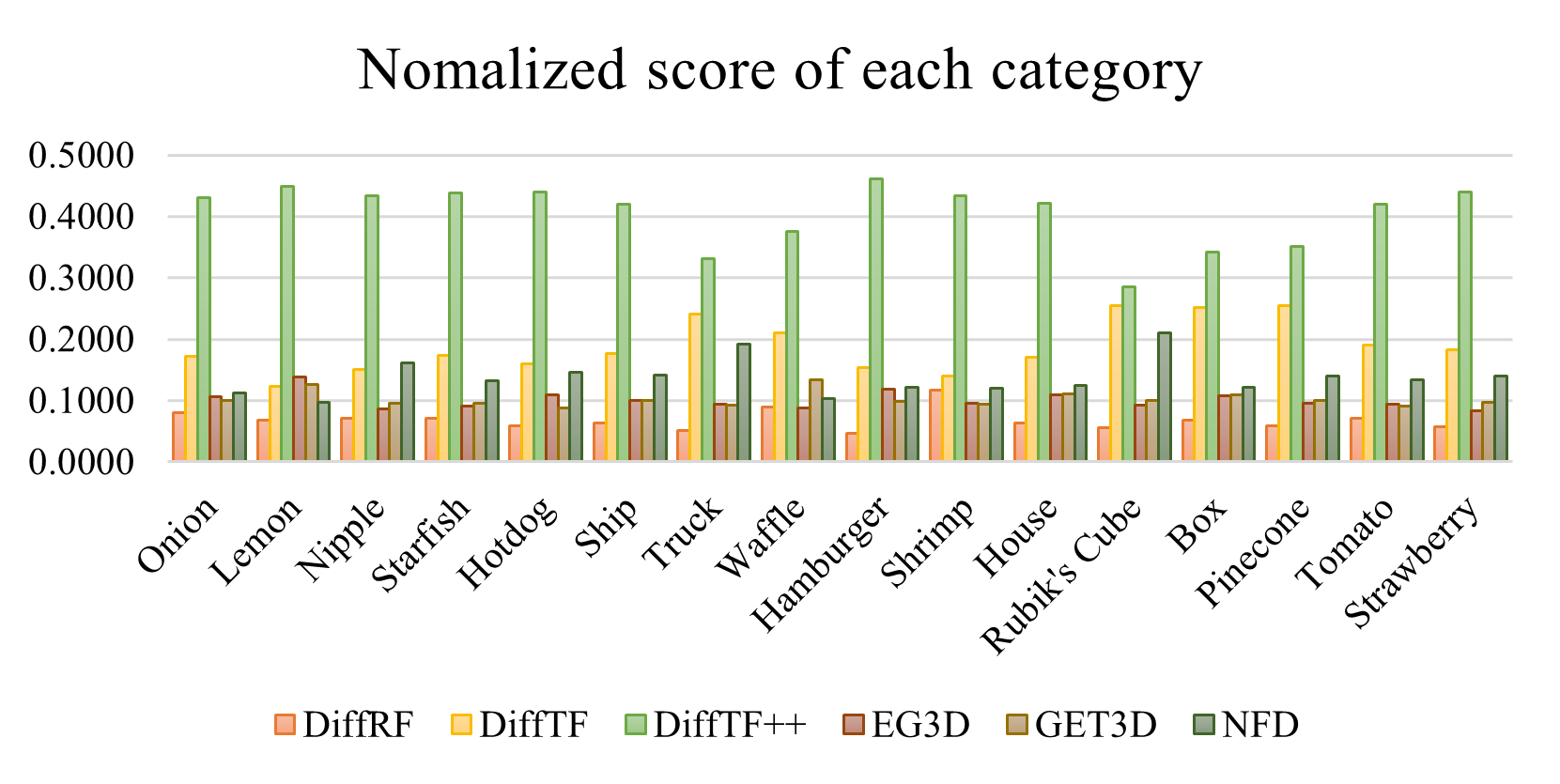}
  \vspace{-20pt}
		\caption{Normalized score of each category on user studies.}\label{user}
	\vspace{-10pt}
\end{figure}
	\subsection{User study}
	To evaluate our methods more comprehensively, we implement user studies. Specifically, we render 360-degree rotating videos for five generative methods~\cite{chan2022efficient,gao2022get3d,muller2022diffrf,shue20223d,cao2023large} and our proposed DiffTF++. There are 96 videos totally in our evaluation, including fruit, truck, vegetable, food, animals, ships, and so on. The volunteers need to rate the performance of generative methods from 1 to 6. For clarification, we average the scores collected from more than 50 people, aged from 24 to 54, including students, staff, and housewives. The overall quantitative results are shown in Fig.~\ref{tab:user}. It demonstrates that our method can generate more diverse and high-quality 3D assets that align with human preferences in large-vocabulary 3D generation. Additionally, we report the normalized score for each category. From Fig.~\ref{user}, we can find that our proposed modules can improve the generative performance compared with DiffTF, especially in categories with detailed surface information, \textit{e.g.}, onion, lemon, shrimp, tomato, and strawberry.

	
	\section{Conclusion}
	In this paper, in contrast to the prior work optimizing the feed-forward generative model on a single category, we propose a diffusion-based feed-forward framework for synthesizing various categories of real-world 3D objects with a single model. To handle the special challenges in large-vocabulary 3D object generation, we 1) improve the efficiency of triplane fitting; 2) introduce the 3D-aware transformer to boost the generative performance in diverse categories; 3) propose the 3D-aware encoder/decoder to handle the categories with complicated topology and texture. Additionally, to further boost the generative performance of our model, we introduce a multi-view reconstruction loss and a neat 3D-aware refinement. By building the connection between the two stages, our DiffTF++ can generate more detailed structures with abundant texture. The exhaustive evaluations against SOTA methods validate the promising performance of our methods. Additionally, user studies demonstrate our generated 3D objects align well with human preferences. We believe that our work can provide valuable insight for the 3D generation community.

	
	

	
	%

	\appendices 
	\section{Evaluation}\label{evaldetail}
 The 2D metrics are calculated between 50k generated images and all available real images. Furthermore, For comparison of the geometrical quality, we sample 2048 points from the surface of 5000 objects and apply the Coverage Score (COV) and Minimum Matching Distance (MMD) using Chamfer Distance (CD) as follows:
	\begin{equation}
		\begin{aligned}
			&CD(X,Y)=\sum\limits_{x\in X} \mathop{min}\limits_{y\in Y}||x-y||^2_2+\sum\limits_{y\in Y} \mathop{min}\limits_{x\in X}||x-y||^2_2,\\
			&COV(S_g,S_r)=\dfrac{|\{\mathrm{\mathop{arg~min}}_{Y \in S_r}CD(X,Y)|X\in S_g\}|}{|S_r|}\\
			&MMD(S_g,S_r)=\dfrac{1}{|S_r|}\sum\limits_{Y\in S_r}\mathop{min}\limits_{X\in S_g}CD(X,Y)
		\end{aligned}
		~ ,
	\end{equation}
	where $X \in S_g$ and $Y \in S_r$ represent the generated shape and reference shape.
	
	Note that we use 5k generated objects $S_g$ and all available real shapes $S_r$ to calculate COV and MMD. For fairness, we normalize all point clouds by centering in the original and recalling the extent to [-1,1]. Coverage Score aims to evaluate the diversity of the generated samples, MMD is used for measuring the quality of the generated samples. 2D metrics are evaluated at a resolution of 128 $\times$ 128. For the Car in ShapeNet, since the GT data contains intern structures, we thus only sample the points from the outer surface of the object for results of all methods and ground truth.

 \textbf{Details about other SOTA methods}
	Since the official NFD merely generates the 3D shape without texture, we reproduce the NFD w/ texture as our baseline. Besides, we use the official code and the same rendering images to train the EG3D and GET3D while the DiffRF and NFD adopt our reproduced code. Note that because of adopting pose condition, official EG3D doesn't support class-conditional generation. Therefore, to maintain the fairness of our evaluations, other class-conditional generative models including DiffTF adopt a random class-conditional input.
 


		
		

		
		

		
		

	\ifCLASSOPTIONcompsoc
	\section*{Acknowledgments}
	\else
	\section*{Acknowledgment}
	\fi
	
	This study is supported by the National Key R\&D Program of China (2022ZD0160201), Shanghai Artificial Intelligence Laboratory, the Ministry of Education, Singapore, under its MOE AcRF Tier 2 (MOE-T2EP20221- 0012), NTU NAP, and under the RIE2020 Industry Alignment Fund – Industry Collaboration Projects (IAF-ICP) Funding Initiative, as well as cash and in-kind contribution from the industry partner(s). $\hfill\blacksquare$
	
	\ifCLASSOPTIONcaptionsoff
	\newpage
	\fi

	\normalem
	\bibliographystyle{IEEEtran}
	\bibliography{journal}
	%
	
	\begin{IEEEbiography}[{\includegraphics[width=1in,height=1.25in,clip,keepaspectratio]{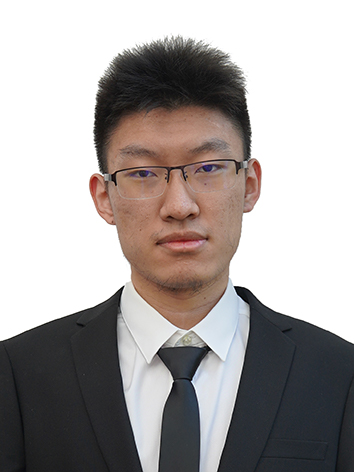}}]{Ziang Cao}
		is currently pursuing a Ph.D. in the School of Computer Science and Engineering at Nanyang Technological University, supervised by Prof. Ziwei Liu. His research interests lie on computer vision, deep learning, and 3D generation.
	\end{IEEEbiography}
	\begin{IEEEbiography}[{\includegraphics[width=1in,height=1.25in,clip,keepaspectratio]{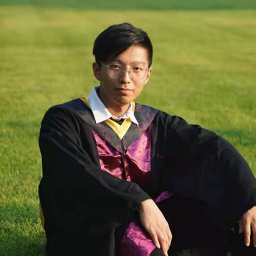}}]{Fangzhou Hong}
		is currently a final-year Ph.D. student in the School of Computer Science and Engineering at Nanyang Technological University (MMLab@NTU), supervised by Prof. Ziwei Liu. Previously, he received B.Eng. degree in Software Engineering from Tsinghua University in 2020. His research interests lie on the computer vision and deep learning. Particularly, he is interested in 3D representation learning and its intersection with computer graphics.

	\end{IEEEbiography}
	\begin{IEEEbiography}[{\includegraphics[width=1in,height=1.25in,clip,keepaspectratio]{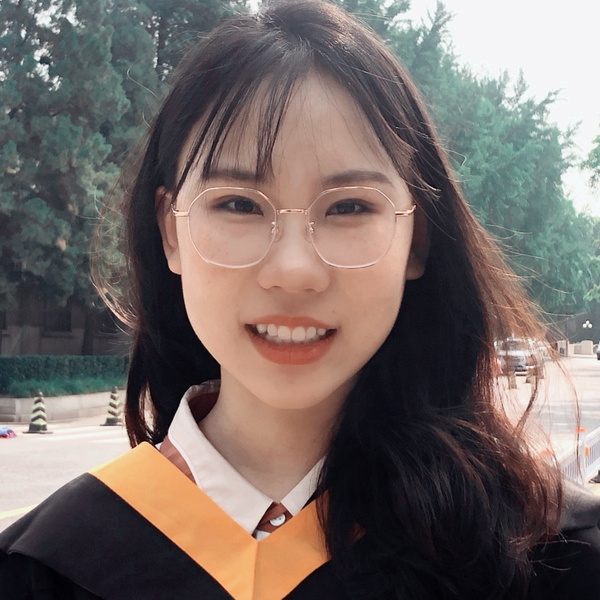}}]{Tong Wu}
		is a final-year Ph.D. student in Multi-Media Lab(MMLab) at CUHK, supervised by Prof. Dahua Lin. She also works closely with Prof. Ziwei Liu at NTU. She is currently a visiting student researcher in Stanford University, working with Prof. Gordon Wetzstein. She got her bachelor’s degree in the EE Department at Tsinghua University in 2020, working with Prof. Yu Wang and Prof. Jiansheng Chen. Her research interests include but not limited to 3D Generation and Reconstruction.
	\end{IEEEbiography}
	\begin{IEEEbiography}[{\includegraphics[width=1in,height=1.25in,clip,keepaspectratio]{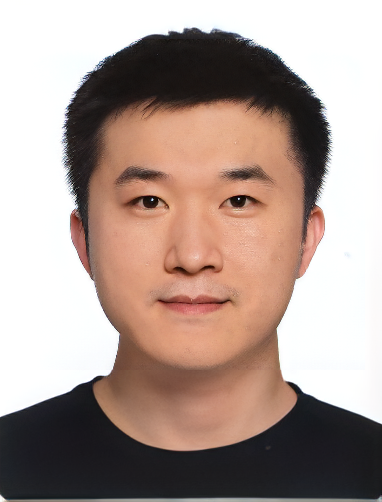}}]{Liang Pan}
		is presently a Researcher at the Shanghai AI Laboratory. He earned his Ph.D. in Mechanical Engineering from the National University of Singapore (NUS) in 2019. He then served as a Research Fellow at the S-Lab of Nanyang Technological University from 2020 to 2023. His research focuses on computer vision, 3D point clouds, and virtual humans. He has made top-tier publications in relevant conferences and journals. Furthermore, he actively contributes to the academic community by serving as a reviewer for esteemed conferences and journals in computer vision, machine learning, and robotics.
	\end{IEEEbiography}
	
	

	\begin{IEEEbiography}[{\includegraphics[width=1in,height=1.25in,clip,keepaspectratio]{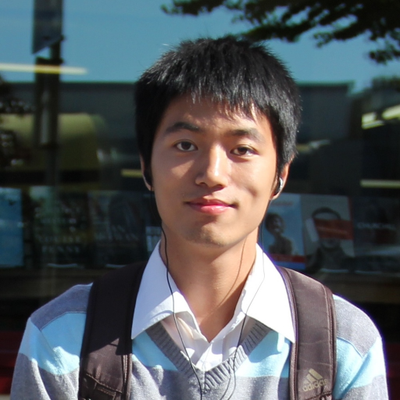}}]{Ziwei Liu}
		Ziwei Liu is currently an Assistant Professor at Nanyang Technological University (NTU). Previously, he was a senior research fellow at the Chinese University of Hong Kong and a postdoctoral researcher at the University of California, Berkeley. Ziwei received his Ph.D. from the Chinese University of Hong Kong in 2017. His research revolves around computer vision/graphics, machine learning, and robotics. He has published extensively on top-tier conferences and journals in relevant fields, including CVPR, ICCV, ECCV, NeurIPS, IROS, SIGGRAPH, TOG, and TPAMI. He is the recipient of the Microsoft Young Fellowship, Hong Kong PhD Fellowship, ICCV Young Researcher Award, and HKSTP best paper award.
		
	\end{IEEEbiography}



	
	

\end{document}